\documentclass{article}

\usepackage{PRIMEarxiv}

\usepackage[utf8]{inputenc} 
\usepackage[T1]{fontenc}    
\usepackage{hyperref}       
\usepackage{url}            
\usepackage{booktabs}       
\usepackage{amsfonts}       
\usepackage{nicefrac}       
\usepackage{microtype}      
\usepackage{lipsum}
\usepackage{fancyhdr}       
\usepackage{graphicx}       
\graphicspath{{media/}}     
\usepackage{amsmath,graphicx,booktabs,amssymb}

\usepackage{subfigure}
\usepackage{algorithm}
\usepackage{algorithmic}
\usepackage{multirow}
\usepackage{caption}
\usepackage{tikz}
\usetikzlibrary{shapes,external}
\usepackage{lipsum}
\tikzset{naming/.style={align=center,font=\footnotesize}}
\tikzset{area/.style = {draw, shape = regular polygon, regular polygon sides = 6, thick, minimum width = 5cm}}

 \usepackage{float}
\usepackage{amsmath}
\usepackage{graphicx}
\usepackage{amssymb}
\usepackage{bm}
\usepackage{comment}
\usepackage{authblk}
\usepackage{multirow}
\usepackage{amsthm}
\usepackage{graphicx}
\usepackage{algorithm}
\usepackage{algorithmic}
\usepackage{url} 
\usepackage{caption}

\usepackage{xcolor}
\usepackage{algorithm}
\usepackage{algorithmic}
\usepackage{hyperref}
 \usepackage{makecell}

\newcommand\tqh[1]{\textcolor{orange}{#1}}

\newcommand\ljn[1]{\textcolor{black}{#1}}

\title{Low-Rank Multitask  Learning based on Tensorized SVMs  and LSSVMs

}

\author{
  Jiani~Liu, Qinghua~Tao, Ce Zhu\thanks{  Corresponding Author. Ce Zhu, Yipeng Liu, and Jiani Liu acknowledge the supports from the National Natural Science Foundation of China (NSFC) under Grant U19A2052, Grant 62020106011 and Grant 62171088. Johan A.K. Suykens and Qinghua Tao acknowledge the supports from iBOF project Tensor Tools for Taming the Curse (3E221427), European Research Council (ERC) Advanced Grant E-DUALITY (787960), KU Leuven 
Grant CoE PFV/10/002, Grant FWO GOA4917N, EU H2020 
ICT-48 Network TAILOR, and Leuven.AI Institute. 
\protect\\
J. Liu, C. Zhu, and Y. Liu are with School of Information and Communication Engineering, University of Electronic Science and Technology of China (UESTC), Chengdu, 611731, China.
Q. Tao, and J.A.K. Suykens are with STADIUS, ESAT, KU Leuven, Heverlee 3001, Belgium.
X. Huang is with the Department of Automation and the MOE Key Laboratory of System Control and Information
Processing, Shanghai Jiao Tong University, Shanghai, 200240, China.
\protect\\
E-mail: jianiliu@std.uestc.edu.cn, qinghua.tao@esat.kuleuven.be., eczhu@uestc.edu.cn, yipengliu@uestc.edu.cn, xiaolinhuang@sjtu.edu.cn, johan.suykens@esat.kuleuven.be. },  Yipeng Liu, Xiaolin Huang, Johan A.K. Suykens }

\begin{document}
\maketitle

\begin{abstract}
 Multitask learning (MTL) leverages task-relatedness to enhance performance. With the emergence of multimodal data, tasks can now be referenced by multiple indices.  In this paper, we employ high-order tensors, with each mode corresponding to a task index, to naturally represent tasks referenced by multiple indices and preserve their structural relations. Based on this representation, we propose a general framework of low-rank MTL  methods with tensorized support vector machines (SVMs) and least square support vector machines (LSSVMs), where the CP factorization is deployed over the coefficient tensor. Our approach allows to model the task relation through a linear combination of shared factors weighted by task-specific factors and is generalized to both classification and regression problems. Through the alternating optimization scheme and the Lagrangian function, each subproblem is transformed into a convex problem, formulated as a quadratic programming or linear system in the dual form. In contrast to previous MTL frameworks, our decision function in the dual induces a weighted kernel function with a task-coupling term characterized by the similarities of the task-specific factors, better revealing the explicit relations across tasks in MTL.  Experimental results validate the effectiveness and superiority of our proposed methods compared to existing state-of-the-art approaches in MTL. The code of implementation will be available at https://github.com/liujiani0216/TSVM-MTL.
\end{abstract}


\section{Introduction}
Multitask learning (MTL) aims to help parameter estimation for each individual task and benefit accurate prediction by exploring the coupling information across different tasks \cite{caruana1997multitask,argyriou2008convex,xu2011multi}. As a promising learning paradigm,  MTL has been widely applied in many fields, such as social sciences\cite{xu2014multi, Evgeniou2004regu, Evgeniou2005learn}, medical diagnosis \cite{han2012implementation, Liang2009Predictive}, 
 etc.   A variety of methods have been developed for MTL and 
 demonstrated promising results in related tasks. 
 {Regarding the techniques} of modelling the task relatedness, representative MTL models can be roughly classified into 3 categories:
 feature-based\cite{argyriou2006multi,argyriou2008convex},  instance-based\cite{zhang2018partially} and parameter-based\cite{chen2009convex,ando2005framework}. 
 
{In this paper, we mainly focus on the parameter-based method, which reflects task relatedness by assuming that related tasks share common model parameters or subspaces. For instance,}
\cite{ando2005framework} assumes that different problems share precisely the same underlying hypothesis space, while \cite{kang2011learning}  first clusters all the tasks into several classes and explores the shared structures of each cluster. \cite{zhu2015subspace} introduces graph Laplacian regularization, which penalizes the coefficients from multiple tasks by preserving sample similarity. In addition, one particularly successful approach is the combination of Multi-Task Learning (MTL) with Support Vector Machines (SVMs) \cite{cortes1995support}. This approach, initially introduced as the Regularized MTL method (rMTL) \cite{Evgeniou2004regu, Evgeniou2005learn}, has shown promising results. As a parameter-based approach, rMTL assumes a shared mean hyperplane across all tasks, while allowing each task's hyperplane to have an offset from this mean hyperplane.  Based on this work, a number of kernel-based methods for MTL have been proposed \cite{lichen2008connection} {and extended into multiclass problems\cite{angulo2003k}.}
Practically, the concept of Least Squares SVM (LSSVM) \cite{suykens1999least} has also been extended to MTL \cite{xu2014multi}, which transforms the inequality constraints in SVMs into equality constraints and simplifies the quadratic programming problem in the dual 
into linear systems.  However, {most current kernel-based MTL models are based on vector/matrix representations, limiting the application to multidimensional datasets or tasks referenced by multiple indices.}


Tensors {are naturally more flexible to provide} effective representations for multidimensional data, {allowing for the preservation of structural information \cite{kolda2009tensor,liu2021tensor,liu2022tensor} {and successful application}
for handling {high-dimensional} \cite{he2017kernelized} data, multiview data\cite{houthuys2021tensor,tao2022tensor,long2023feature} and complex functions \cite{chen2017parallelized} in single task learning.}
In recent years, different MTL models have been proposed by utilizing tensor representations \cite{zheng2019multitask,xu2019spatio,zhang2019tensor,yang2017deep,zhang2020deep,romera2013multilinear,wimalawarne2014multitask}, {and  mainly consist of two {groups.}
The {former}
focuses on handling multidimensional inputs for each task \cite{zheng2019multitask,xu2019spatio,zhang2019tensor}; these approaches leverage the inherent structure of the input tensors to improve MTL performance.
{The latter deals} with tasks referenced by multiple indices \cite{romera2013multilinear,wimalawarne2014multitask,zhao2019multilinear,zhang2023multilinear}. Specially, in real scenarios,  the involved tasks are relevant and referenced by multiple indices according to physical meanings. For example, in disease progression prediction, learning problem according to different clinical {criteria}
and time points can be constructed. Instead of referencing all the tasks using a single index, {i.e., $t=1, \ldots, T$}, it is more natural to follow the physical meaning of how we group these tasks in practice, which means letting one index $t_1=1,\cdots,T_1$ for ``clinical criterion", and the other $t_2=1,\cdots,T_2$ for ``time point", to indicate those {$T_1\times T_2=T_1T_2 \triangleq T$} tasks.


Drawing on the tensor representation of the tasks with multiple indices, \cite{romera2013multilinear} proposes to impose a low Tucker \cite{Tuck1963a,levin1965three} rank regularization on the stacked coefficient tensor, {allowing for the exploration of task relations in MTL.}
The proposed approach encompasses two optimization strategies: rank minimization and tensor factorization.  Subsequently, in order to extend the applicability of rank minimization approaches to large datasets, various convex surrogates for Tucker rank are proposed\cite{wimalawarne2014multitask,zhao2019multilinear}. {In \cite{signoretto2013multilinear}, it generalizes the MLMTL problem with learning tensor-based models in Reproducing kernel Hilbert spaces with multilinear spectral penalties, allowing for nonlinear extensions.} Additionally, a new multilinear multitask learning (MLMTL) method is proposed recently by leveraging the transformed tensor nuclear norm \cite{zhang2023multilinear}, which explores the low multi-rank structure in a transformed tensor. However, all these models are based on the Tucker decomposition or tensor singular value decomposition (t-SVD)\cite{kilmer2011factorization,kilmer2008third}. Tucker suffers from a dimensionality curse if the tensor order increases, while t-SVD is only suitable for third-order tensors and sensitive to order permutation. 
It is worth noting that these approaches primarily focused on multilinear modeling. However, it is widely acknowledged that nonlinearity plays a crucial role in effectively capturing {more informative patterns} 
of both {complicated} data and tasks. Therefore, incorporating nonlinearities into the models is highly desirable for achieving more accurate and comprehensive representations {in MTL}.

{Motivated by these, we develop a novel MTL approach based on tensorized LSSVMs upon our previous work \cite{liu2023tensorized}, only covering the regression tasks with experimental results. 
Towards a more generic methodology, we develop a unified optimization framework of low rank multitask learning methods based on the tensorization of both SVMs and LSSVMs in classification and also regression tasks.} {In addition, more comprehensive experimental results and further analysis regarding the convergence, computation efficiency, and the kernel-based decision functions are given for completeness.} {The main contribution of this paper is summarized as follows.}
\begin{itemize}
     \item In this paper, we propose a unified MTL framework 
     {leveraging} Tensorized SVMs \ljn{and LSSVMs}
     {on} both regression and classification problems.  It constructs a high-order weight tensor by grouping the involved tasks with multiple indices, aligning with their respective physical interpretations. In addition, both linear and nonlinear feature maps (or kernels) can be employed, providing flexibility in modeling.
    \item Instead of using Tucker or t-SVD in previous work, we factorize the constructed tensor into CP forms. This allows us to have a simple representation while still enabling tasks to share a common latent factor and maintain task-specific factors. In addition, an 
    illustration of how the tensor representation reflects the task relations through the CP factorization is also provided.
    \item  Based on alternating optimization concept and the Lagrangian {function,}
    each subproblem successfully degenerates to be convex 
    as a quadratic programming or linear system.
    The decision function derived in the dual 
    reveals a task-coupling term that introduces weighted kernel functions, 
    capturing the relationship between different tasks.

\end{itemize}
 Numerical experiments on simulated data, {real-world benchmark data,}
 and also the application to Disease progression prediction task show  advantageous performances of proposed approaches over the matrix-based and other existing tensor-based MTL methods. 


The remainder of this paper is organized as follows.
Section \ref{sec: notations}  gives some  premieres. Section \ref{sec: method} presents the  modelling and optimization for our tLSSVM-MTL. Experimental results and conclusions are  in Sections \ref{sec: experiment} and \ref{sec: conclusion}. 

\section{Notations and Preliminaries}
\label{sec: notations}
\subsection{Notations}
Scalars, vectors, matrices and tensors are represented as $a$, $\mathbf{a}$, $\mathbf{A}$ and $\mathcal{A}$. {For an} $N$-th order tensor {represented as $\mathcal{A}\in\mathbb{R}^{I_1\times \cdots\times I_N}$, it is an} $N$-way array with $\prod_{n=1}^N I_n$ elements, {where the} $i_1,\cdots,i_N$-th entry is denoted as $\mathcal{A}(i_1,\cdots,i_N)$ or $a_{i_1,\cdots,i_N}$. In this way, scalars, vectors, matrices can also be seen as $0$-th, $1$-th, $2$-th order tensors, respectively.  For clarity, we specially denote the row vectors of matrix $\mathbf{A}\in\mathbb{R}^{I\times J}$ as $\mathbf{A}[i,:]^T = \mathbf{a}_{i,:}$, its column vectors as $ \mathbf{A}[:, j] = \mathbf{a}_{:,j}$. The outer product of vector $\mathbf{a}\in\mathbb{R}^{I}$ and $\mathbf{b}\in\mathbb{R}^{J}$ yields a matrix 
$\mathbf{C}\in\mathbb{R}^{I\times J}$.  The cardinality of a set $\mathbb{S}$ is denoted as $\lvert \mathbb{S}\lvert$. For matrix $\mathbf{A}\in\mathbb{R}^{I\times I}$, $\operatorname{tr}(\mathbf{A})=\sum_{i} a_{i,i}$ represents the trace operator.  {The multi-index $i=\overline{i_1,\cdots,i_N}$ either stands for $\overline{i_1,\cdots,i_N}=i_1+(i_2-1)I_1+\cdots+(i_N-1)I_1I_2\cdots I_{N-1}$ in little-endian convention or $\overline{i_1,\cdots,i_N}=i_N+(i_{N-1}-1)I_N+\cdots+(i_1-1)I_2 \cdots I_{N}$ in little-endian. }

\noindent \textbf{Definition: CP decomposition} The CP decomposition of a given $N$-th order tensor $\mathcal{X}\in\mathbb{R}^{I_1\times \cdots\times I_N}$ is expressed by
\begin{equation}
    \mathcal{X}=\sum_{r=1}^{R} \mathbf{u}^{1}_r \circ \cdots \circ \mathbf{u}^{N}_r
\end{equation}
which factorizes the tensor $\mathcal{X}$ into a summation of several rank one components \cite{carroll1970analysis,harshman1970foundations}, {where $\mathbf{u}_r^n\in\mathbb{R}^{I_n}$ for all $n=1,\cdots,N$, $r=1,\cdots, R$.} It can also be {element-wisely} represented in matrix form as follows
\begin{equation}
    \mathcal{X}(i_1,\cdots,i_N)=  \sum_{r=1}^R \mathbf{U}^1(i_1,r)   \cdots  \mathbf{U}^{N}(i_N,r)
\end{equation}
where $\mathbf{U}^n=[\mathbf{u}
^n_1,\cdots,\mathbf{u}^n_R]$, $n=1,\cdots,N$ are factor matrices. The smallest number of rank one components required to represent the given tensor $\mathcal{X}$ in this CP format  is denoted as CP rank. In the following part, we represent the CP decomposition as $\mathcal{X}=[\![\mathbf{U}^{1},\cdots,\mathbf{U}^{N} ]\!]$ for simplicity. Fig. \ref{fig: cp} gives a graphical illustration of the CP decomposition of a given third order tensor.
 
  \begin{figure}[t]
    \centering
\includegraphics[scale=0.5]{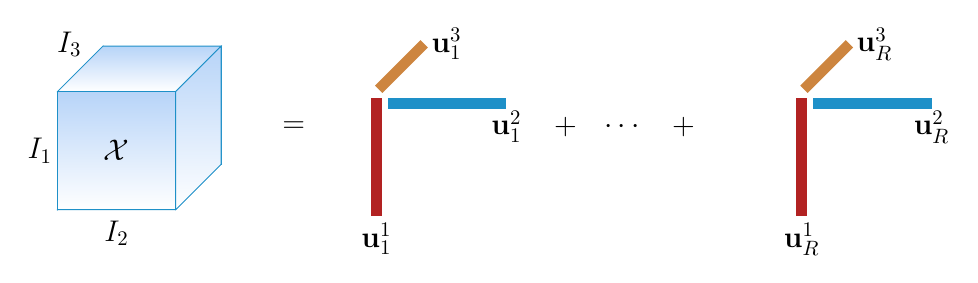}
 \caption{CP decomposition for a third order tensor.}
    \label{fig: cp}
\end{figure}

\subsection{Support Vector Machine for Multitask Learning}
{Without loss of generality, we consider $T$  tasks in the addressed MTL.}
For each task $t$, we have a training dataset $\mathcal{D}_{t}=\{\mathbf{x}_{i}^{t}\in\mathbb{R}^{d_{t}},y_{i}^{t}\}_{i=1}^{m_{t}}$,
with $y_i^t\in\{1,-1\}$ for classification and $y_i^t\in\mathbb{R}$ for regression. The $i$-th predictor and response of $t$-th task is $\mathbf{x}_i^t$ and $\mathbf{y}_i^t$, respectively. The number of training samples for $t$-th task is $m_{t}$ {and then the total} sample size is $m=\sum_{t=1}^{T} m_t$. {Here we 
focus on homogeneous features in MTL, 
implying}
$d_{t}=d$ for $t=1,\cdots, T$. 

Based on hierarchical Bayesian methods \cite{allenby1998marketing,arora1998hierarchical}, RMTL \cite{Evgeniou2004regu,Evgeniou2005learn} extends the kernel-based method SVM for multitask learning. RMTL assumes that all tasks share a common mean hyperplane, with each task's hyperplane having an offset from the mean hyperplane. This allows for capturing the shared information among tasks while accommodating task-specific variations. Specifically, the model weights $\mathbf{w}_t$ satisfy $\mathbf{w}_t=\mathbf{w}_0+\mathbf{v}_t$, where $\mathbf{w}_0$ is a common part for all tasks while $\mathbf{v}_t$ carries the task-specific information. 

RMTL finds the coefficients $\mathbf{w}_0\in\mathbb{R}^{d_h},\{\mathbf{v}_t\in\mathbb{R}^{d_h}\}_{t=1}^T$ for the classification problem through the following optimization problem
\begin{eqnarray}
\underset{\mathbf{w}_0,\mathbf{v}_t, \xi_{i}^{t}}{\arg \min } & 
\sum_{t=1}^{T} \sum_{i=1}^{m_{t}}\xi_{i}^{t}+\frac{\lambda_1}{T}\sum_{t=1}^{T} \lVert\mathbf{v}_{t} \lVert^2 +\lambda_2  \lVert\mathbf{w}_{0} \lVert^2 \nonumber\\ \nonumber
\text { s.t. } & y_{i}^{t}\left( (\mathbf{w}_0+\mathbf{v}_t)^{\top} \phi(\mathbf{x}_{i}^{t})\right)\geq1-\xi_{i}^{t}
;  { \xi_{i}^{t} \geq 0 }
\nonumber
\end{eqnarray}
with $i=1, \ldots, m_{t}$, $t=1, \ldots, T$, $\lambda_1,\lambda_2 \geq 0$ are 
regularized parameters, $\phi(\cdot): \mathbb{R}^{d}\mapsto\mathbb{R}^{d_h}$ is a linear or nonlinear function {mapping the inputs to a feature space which can be very high dimensional.}
Through a novel designed kernel function with a task-coupling parameter, the optimization problem can be formulated as a standard SVM problem and solved correspondingly. 


\section{\ljn{Tensorized SVMs and LSSVMs for MTL}}
\label{sec: method} 
\subsection{Tensor representation for Multitask learning}
\label{sec:w:tensor}
{In real-world scenarios of MTL, it is common that the tasks  can be characterized into different types that exert explicit} physical meanings.
For example, in  restaurant rating systems, the predictions {consists of tasks that involve} 2 aspects, i.e., ``consumers''  and ``rating aspect'' (food, service, environment).  Rather than using a single index $t=1, \ldots, T$ for all involved tasks in traditional MTL settings, it is more natural to 
{model these tasks with} 
two indices $t=\overline{t_1 t_2}$, $t_1=1, T_1$ for ``consumer'' and $t_2=1, 2, 3$ for ``rating aspect'', {as illustrated in Fig. \ref{fig:mprrs}.} 
For each consumer, the restaurant rating system aims to predict different aspects (food, service, environment) of restaurants based on their attributes  such as location, decoration, etc. Considering the $t_1$-th consumer and $t_2$-th rating aspect, the goal is to learn function $f_{t_1,t_2}$ from dataset $\{\mathbf{x}_i^{t_1,t_2}\in\mathbb{R}^{d},y_i^{t_1,t_2}\}_{i=1}^{m_{t_1,t_2}}$, where $\mathbf{x}_i^{t_1,t_2}$ and $y_i^{i_1,t_2}$ are {the} $i$-th pair sample for the $t_1,t_2$-th task, and $m_{t_1,t_2}$ is the number of samples for the $t_1,t_2$-th task. {Regarding the essence of MTL,}
the exploration of high-order task-relatedness 
has  {thereby} become a key issue \cite{romera2013multilinear}.

  \begin{figure}[t]
    \centering
\includegraphics[scale=0.8]{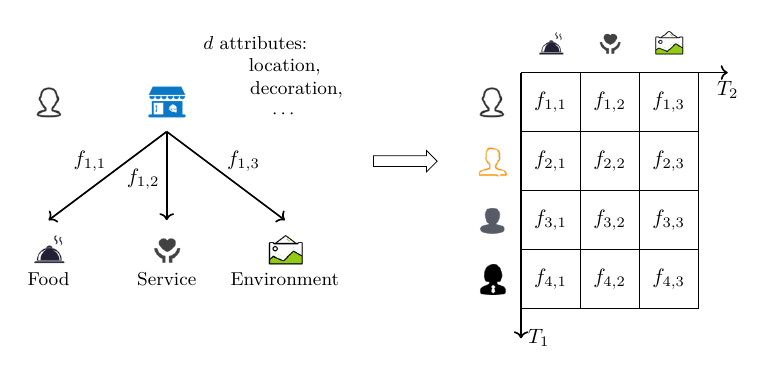}
\caption{{An illustrative example of} Restaurant rating system {with multiple indices for tasks.}}
\label{fig:mprrs}
\end{figure}

{With {the} kernel-based learning framework leveraged in this work, i.e., SVMs,} we assume that the function $f_{t_1,t_2}$ for the $t_1,t_2$-th task is 
{given by} $f_{t_1,t_2}(\mathbf{x})={\mathbf{w}_{t_1,t_2}^{\top}} \phi(\mathbf{x}) +b^{t_1,t_2}$, where $\mathbf{w}_{t_1,t_2}\in\mathbb{R}^{d_h}$, $b_{t_1,t_2}$ are  the {task-specific} weight vector and the bias.
Collecting all the weight vectors together, we can naturally get a tensor $\mathcal{W}\in \mathbb R^{d_h \times T_1 \times T_2}$, where {the} first mode $d_h$ denotes the number of features,  {the} second mode $T_1$ denotes the number of consumers, {the} third mode $T_2$ denotes the number of rating aspects, as shown in Fig. \ref{fig: wt-to-w}. 

\begin{figure}[t]
    \centering
\includegraphics[scale=1.8]{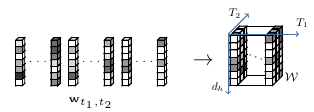}
    \caption{Tensorized representation of weight vectors in kernel-based learning, {on the illustrative example in Fig. \ref{fig:mprrs}}.}
    \label{fig: wt-to-w}
\end{figure}

In order to mine the higher-order task-relatedness, we apply the low-rank CP decomposition over the weight tensor $\mathcal{W}$ as follows
\begin{equation}\label{eq:tensor:w:cp}
\mathcal{W}=\sum\nolimits_{r=1}^R \mathbf{l}_{:,r}\circ \mathbf{u}^{1}_{:,r}\circ \mathbf{u}^{2}_{:,r}=[\![ \mathbf{L},\mathbf{U}^{1},\mathbf{U}^{2}]\!],
\end{equation}
where $\mathbf{L}=[\mathbf{l}_{:,1}; \cdots; \mathbf{l}_{:,R}]\in\mathbb{R}^{d_h\times R}$ is the shared {factor} exploiting the coupling information across all tasks, $\mathbf{U}^1=[\mathbf{u}^1_{:,1}, \ldots, \mathbf{u}^1_{:,R} ]  \in\mathbb{R}^{T_1\times R}$ is the factor corresponding to the index for consumers, and $\mathbf{U}^2=[\mathbf{u}^2_{:,1}, \ldots, \mathbf{u}^2_{:,R} ]  \in\mathbb{R}^{T_2\times R}$ is the factor corresponding to the index for rating aspects. Then the weight vector of the $t_1,t_2$-th task, namely $\mathbf{w}_{t_1,t_2}$, can be correspondingly represented as 
\begin{equation}\label{eq:tensor:w:vector}
\mathbf{w}_{t_1,t_2}=\sum\nolimits_{r=1}^R \mathbf{l}_{:,r}\cdot {u}^{1}_{t_1,r}  {u}^{2}_{t_2,r}.
\end{equation} 
As shown in \eqref{eq:tensor:w:vector}, the weight vector can be seen as a linear combination of $R$ factors $\{\mathbf{l}_{:,r}\}_{r=1}^R$ characterized by the weights $\{u_{t_1,r}^1, u_{t_2,r}^2\}_{r=1}^R$. For all the involved tasks, $\{\mathbf{l}_{:,r}\}_{r=1}^R$ are shared and learns the coupling information across tasks. {In contrast, $\{u_{t_1,r}^1, u_{t_2,r}^2\}_{r=1}^R$ vary with different tasks and mine the task relatedness.} In this way, the modelling parameters for all tasks are coupled in a multidimensional space, {as exemplified in Fig. \ref{fig: wt-cp}.}

\begin{figure}[t]
    \centering
\includegraphics[scale=1.5]{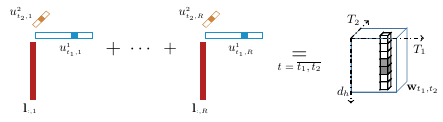}
    \caption{An 
    illustration for weight vector $\mathbf{w}_{t_1,t_2}$ of $t_1,t_2$-th task under low rank CP decomposition, {on the illustrative example in Fig. \ref{fig:mprrs}}.}
    \label{fig: wt-cp}
\end{figure}

Particularly, it is worth noting that the 
{applied tensor representation 
enables
to 
capture  dependencies 
using multiple modes than 
a single mode.} Specifically, 
the update of $\{u_{t_1,r}^1\}_{r=1}^R$ 
{is associated with}
the tasks 
{regarding} the $t_1$-th consumer while the update of $\{u_{t_2,r}^2\}_{r=1}^R$ only affects the tasks related to the $t_2$-th rating aspect. In other words, all the samples related to the $t_1$-th consumer will be considered when updating $\{u_{t_1,r}^1\}_{r=1}^R$, seeking personal preference trends of $t_1$-th consumer;
all the samples related to the $t_2$-th rating aspect will be considered when updating $\{u_{t_2,r}^2\}_{r=1}^R$, exploring the common characteristics of different users when rating the $t_2$-th aspect. In other words, $\{u_{t_1,r}^1\}_{r=1}^R$ explores the task-relatedness along the first mode, which corresponds to different consumers, while $\{u_{t_2,r}^2\}_{r=1}^R$ explores the task-relatedness along the second mode, \tqh{} corresponding to a different rating aspect with the first model.}

 More generally, we consider the multitask learning problem with $N$ indices, i.e, $T_1,\cdots,T_N$. {Hence,} the constructed tensor is $\mathcal{W}\in\mathbb{R}^{d_h\times T_1 \times \cdots\times T_N}$
 factorized as
\begin{equation} 
\mathcal{W}=\sum\nolimits_{r=1}^R \mathbf{l}_{:,r}\circ \mathbf{u}^{1}_{:,r}\circ \cdots\circ \mathbf{u}^{N}_{:,r}=[\![ \mathbf{L},\mathbf{U}^{1},\ldots,\mathbf{U}^{N}]\!],
\end{equation}
 where $\mathbf{U}^n\in\mathbb{R}^{T_n\times R}$ for $n=1,\cdots,N$ and the total number of tasks is $T = \prod_{n=1}^N T_n$.  The CP rank $R$ indicates the number of latent shared features $\mathbf{l}_{:,r}$ in this representation.
 




\subsection{Tensorized SVMs for MTL}
\label{sec: tsvm}
{With the constructed} tensor representation for 
{MTL tasks referenced by}  multiple indices, we 
can formulate the corresponding optimization problem 
{under} the SVM framework for both classification and regression problems.
For clarity, ${t}$ is used 
to denote
the $\overline{t_1,\cdots,t_N}$-th task. 
\subsubsection{Classification}
First, we consider the tensorized SVMs for classification, i.e., tSVC-MTL.  The primal optimization problem with the deployed low rank CP factorization is formulated as

\begin{eqnarray}\label{eq:cla-TSVM-MTL:primal}
\underset{\Theta}{ \min } & 
{{C}} \sum_{t} \sum_{i}\xi_{i}^{t}+\frac{1}{2}\operatorname{tr} \mathbf{L L}^{\top}+\frac{1}{2}\sum_{n} \operatorname{tr} \mathbf{U}^{n}  {\mathbf{U}^{n}}^{\top}\\ \nonumber
\text { s.t. } & y_{i}^{t}\left( ( \sum_{r} \mathbf{l}_{:,r}  {u}_{t_{1},r}^{1}  \cdots {u}_{t_{N},r}^{N}  )^{\top} \phi(\mathbf{x}_{i}^{t})+b_{t}\right)\geq 1-\xi_{i}^{t}, \nonumber\\
& \xi_{i}^{t} \geq 0,  t=\overline{t_{1},\cdots,t_{N}}  \nonumber.
\end{eqnarray}
where $\Theta=\{\mathbf{L}, \mathbf{U}^{n}, b_{t}\}$ is the collection of unknown variables, $t=1,\cdots,T$, $i=1,\cdots,m_t$, $r=1,\cdots,R$. {In \eqref{eq:cla-TSVM-MTL:primal}, 
SVMs is extended to simultaneously deal with different} tasks referenced by multiple indices. {The low-rank CP  factorization {is succinct in representation for the coefficient in MTL and is able}  to explicitly {decouple}
the shared factor $\mathbf{L}$ 
{exploring} common information {across all tasks;  in the meantime the factor $\mathbf{U}^n$ mining} task relatedness along $n$-th index/mode of involved  tasks,} together boosting the overall performance of all tasks in MTL. 
 
As shown in 
\eqref{eq:cla-TSVM-MTL:primal}, the product operations between the shared factor $\mathbf{L}$ and the task-specific factors $\{\mathbf{U}^1,\ldots, \mathbf{U}^N\}$ of the coefficient tensor
{lead to a nonconvex problem in optimization.}
In order to solve this problem, we  update each latent factor while the others are fixed based on the alternating minimization technique,
i.e., the block coordinate descent.

\noindent {\bf{ 1) {$(\mathbf{L}, b_t, \xi_i^t)$}-subproblem}}
The training objective function w.r.t. $\mathbf{L}, b_t, \xi_i^t$ {follows from the Lagrangian }
\begin{eqnarray}\label{eq: cla-l-obj}
\mathcal{L}(\mathbf{L}, b_t, \xi_i^t)={\boldsymbol{C}} \sum_{t} \sum_{i}\xi_{i}^{t}+\frac{1}{2}\operatorname{tr} \mathbf{L L}^{\top}- \sum_{t}\sum_{i}\alpha_{i}^{t}
[ y_{i}^{t}(\mathbf{u}_{t}^{\top} \mathbf{L} ^{\top}\phi(\mathbf{x}_{i}^{t})+b_{t})-1+\xi_{i}^{t} ]-\sum_{t}\sum_{i} \gamma_i^t \xi_i^t
\end{eqnarray}
where $\mathbf{u}_{t} \triangleq [u_{t,1}, \ldots, u_{t, R}]^{\top}\in\mathbb{R}^{R}$, $u_{t,r}\triangleq {u}_{t_{1},r}^{1}  \cdots {u}_{t_{N},r}^{N}$ for $t=\overline{t_{1},\ldots,t_{N}}, t_{n}=1,\ldots,T_{n}, n=1, \ldots, N$, $i=1,\ldots, m_t$. $\alpha_i^t$ and $\gamma_i^t$ are Lagrangian multipliers.
 
{Then the stationary point conditions are obtained as}
 \[\arraycolsep=1.4pt\def\arraystretch{2.2}
\begin{array}{l}
  \dfrac{\partial \mathcal{L}}{\partial \mathbf{L}}=0 \Longrightarrow \mathbf{L}=\sum_{t} \sum_{i} \alpha_{i}^{t} y_{i}^{t} \phi(\mathbf{x}_{i}^{t}) \mathbf{u}_{t}^{\top}, \nonumber\\ 
\dfrac{\partial \mathcal{L}}{\partial {b}_t}=0\Longrightarrow \sum_{i=1}^{m_t} \alpha_i^ty_i^t=0, \nonumber  \\
\dfrac{\partial \mathcal{L}}{\partial \xi_i^t}=0 \Longrightarrow   C  ={\alpha_i^t}+  \gamma_i^t.\nonumber   
\end{array}
\]
  
The dual problem can then be expressed as
\begin{eqnarray}\label{eq: cla-l-sol}
{\max_{ \alpha_i^t}} &{-}\frac{1}{2} \sum_{t} \sum_{q}\sum_{i}\sum_{p}\alpha_i^t\alpha_p^q y_i^t y_p^q \langle\mathbf{u}_t,\mathbf{u}_q\rangle k(x_i^t,x_p^q) {+}\sum_{t} \sum_{i}\alpha_i^t \\
\text { s.t. } &  \sum_{i=1}^{m_t} \alpha_i^ty_i^t=0, 0\leq \alpha_i^t \leq C, 
\nonumber
\end{eqnarray}
where $\alpha_i^t$ can be obtained by solving this convex QP problem, $t,q=1,\cdots,T$, $i=1,\cdots,m_t$, $p=1,\cdots,m_q$.

\noindent {\bf{ 2)  {$(\mathbf{U}^{n}, b_t, \xi_i^t)$}-subproblem}} 
{Acccordingly,} the update of factor $\mathbf{U}^n\in\mathbb{R}^{T_n\times R}$ can be divided into $T_n$ independent subproblems along the first mode.  The training objective function w.r.t. $\mathbf{U}^{n}_{t_n,:}\in\mathbb{R}^{R}$ {follows from the Lagrangian }
 \begin{eqnarray}\label{eq: cla-u-obj}
\mathcal{L}(\mathbf{u}^{n}_{t_{n},:}, b_{t}, \xi_i^t)= {{C}} \sum_{t\in\mathbb{O}^n_{t_n}} \sum_{i}\xi_{i}^{t}+\frac{1}{2}\lVert \mathbf{u}_{t_{n},:}^{n}\lVert_{2}^2 -\sum_{t\in\mathbb{O}^n_{t_n}} 
\sum_{i}\beta_{i}^{t}[y_{i}^{t}\left(  {\mathbf{u}_{t_{n},:}^{n} }^{\top}\mathbf{z}_{i}^{t} +b_{t}\right)- 1+\xi_{i}^{t}] -\sum_{t\in\mathbb{O}^n_{t_n}} \sum_{i}\eta_i^t \xi_{i}^{t} 
\end{eqnarray}
where $\mathbf{z}_{i}^{t}$ is  formulated as $\mathbf{L}^{\top}  {\phi(\mathbf{x}_{i}^{t}) } \odot \mathbf{u}_{t_{1},:}^1\odot\cdots\odot \mathbf{u}_{t_{n-1},:}^{n-1}\odot\mathbf{u}_{t_{n+1},:}^{n+1}\odot\cdots\odot \mathbf{u}_{t_{N},:}^N\in\mathbb{R}^R$,  the index set containing the involved tasks is given by
$\mathbb{O}^n_{t_n}=\{\overline{t_{1}, \ldots, t_{N}}| t_{l}=1,\ldots,T_{l}, l=1, \ldots, N, l \neq n\}$ with cardinality
$|\mathbb{O}^n_{t_n}|=\prod_{l,l\neq n}T_l$, and the indices in constraints satisfy $t\in\mathbb{O}^n_{t_n}, i=1, \ldots, m_t$.  Here the total number of constraints is ${M^n_{t_n}}=\sum_{t\in \mathbb{O}^n_{t_n}} m_t$.  $\beta_i^t$ and $\eta_i^t$ are Lagrangian multipliers.

Finding the stationary points we get 
  \[\arraycolsep=1.4pt\def\arraystretch{2.2}
\begin{array}{rll}\dfrac{\partial \mathcal{L}}{\partial \mathbf{u}^{n}_{t_{n},:}}=0 &\Longrightarrow&\mathbf{u}^{n}_{t_{n},:}=\sum_{t\in\mathbb{O}^n_{t_n}}\sum_{i=1}^{m_t} \beta_{i}^{t} y_{i}^{t} \mathbf{z}_{i}^{t}, \nonumber\\ 
\dfrac{\partial \mathcal{L}}{\partial b_t}=0 &\Longrightarrow & \sum_{i=1}^{m_t} y_i^t \beta_i^t=0, \nonumber\\
\dfrac{\partial \mathcal{L}}{\partial \xi_i^t}=0  &\Longrightarrow&  C  ={\beta_i^t}+  \eta_i^t. \nonumber 
\end{array}\] 
 
The dual problem can be expressed as
\begin{eqnarray}\label{eq: cla-u-sol}
{\max_{\bm\beta^t_{i}}} &{-}\frac{1}{2} \sum_{t\in\mathbb{O}_{t_n}^n}  \sum_{q\in\mathbb{O}_{t_n}^n}\sum_{i}\sum_{p} \beta_i^t\beta_p^q y_i^t y_p^q \langle\mathbf{z}_i^t,\mathbf{z}^q_p\rangle
{+}\sum_{t\in\mathbb{O}_{t_n}^n} \sum_{i} \beta_i^t \\
\text { s.t. } & \sum_{i} \beta_i^ty_i^t=0;  \nonumber\\
& 0\leq \beta_i^t \leq C  \text{ for } t\in\mathbb{O}_{t_n}^n   \nonumber
\end{eqnarray}
where $\beta_i^t,t\in\mathbb{O}_{t_n}^n$ can be obtained by solving this convex QP problem, $ i=1,\cdots, m_t$, $ p=1,\cdots, m_q$. A more detailed derivation process can refer to Appendix section \ref{appendix 1-cla}. The updating steps are summarized in Algorithm \ref{algorithm:tSVM-MTL}. 

\begin{algorithm}
	\caption{tSVC-MTL}
	\label{algorithm:tSVM-MTL}
	\begin{algorithmic}
		\STATE \textbf{Input}: ${y^t_i},x^t_i$ for $t=1,\cdots,T$ $i=1,\cdots,m_t$
		\STATE \textbf{Output}: dual variables ${\alpha_i^t}$ for $i=1,\cdots,m_t$, $t=1,\cdots,T$, $b_t$ for $t=1,\cdots,T$, $\mathbf{U}^n$ for $n=1,\cdots,N$.
		\STATE $\bullet$ Initialize $\mathbf{U}^n$ for $n=1,\cdots,N$
		\WHILE{\textbf{not convergence}}
		\STATE $\bullet$ update ${\alpha_i^t}, b_t$ by solving  the convex QP problem (\ref{eq: cla-l-sol})		
		\FOR{$n=1:N$}
		    \FOR{$t_n=1:T_n$}
		         \STATE $\bullet$ update $\{\beta_i^t, b_t, t\in\mathbb{O}_{t_n}^n, i=1,\cdots, m_t\}$ by solving the convex QP problem (\ref{eq: cla-u-sol})
		       \STATE $\bullet$ $ \mathbf{u}^{n}_{t_{n},:}=\sum_{t\in\mathbb{O}^n_{t_n}}\sum_{i=1}^{m_t} \beta_{i}^{t} y_{i}^{t} \mathbf{z}_{i}^{t} $ 
		    \ENDFOR
		\ENDFOR
		\ENDWHILE
	\end{algorithmic}
\end{algorithm}

\subsubsection{Regression}
For regression, the optimization problem of
our tensorized SVMs,
i.e., tSVR-MTL, is constructed in the primal as: 
\begin{eqnarray}\label{eq:reg-tsvm-mtl:primal}
\underset{\Theta}{\min } &  {C} \sum_{t}\sum_{i}(\xi_{i}^{t}+{\xi_i^t}^*) +\frac{1}{2}\operatorname{tr} \mathbf{L L}^{\top}+\frac{1}{2}\sum_{n} \operatorname{tr} \mathbf{U}^{n}  {\mathbf{U}^{n}}^{\top} 
\quad\\ 
\text { s.t. } &
f(\mathbf{x}_i^t)=( \sum\nolimits_{k} \mathbf{l}_{:,k}  \cdot {u}_{t_{1},k}^{1}  \ldots {u}_{t_{N},k}^{N}  )^{\top} \phi(\mathbf{x}_{i}^{t})+b_{t} \nonumber\\
& y_{i}^{t}-f(\mathbf{x}_i^t)\leq \xi_{i}^{t}+\varepsilon ,\nonumber \\
& f(\mathbf{x}_i^t)-y_{i}^{t}\leq {\xi_{i}^{t}}^*+\varepsilon ,\nonumber\\
& \xi_{i}^{t}, {\xi_i^t}^*\geq 0,\quad t=\overline{t_1,\ldots,t_N}. \nonumber
\end{eqnarray}
where $\Theta=\{\mathbf{L}, \{\mathbf{U}^{n}\}_{n=1}^N, \{b_{t}\}_{t=1}^T\}$ is the collection of unknown variables to be determined.

\noindent 1) {\bf{{$(\mathbf{L},b_t,\xi_i^t,{\xi_i^t}^*)$}-subproblem}} 
 
The {objective} function w.r.t. 
$\mathbf{L},b_t,\xi_i^t,{\xi_i^t}^*$ {follows from the Lagrangian } 
\begin{eqnarray}
&\mathcal{L}({\mathbf{L},  b^{t},\xi_i^t,{\xi_i^t}^*})=  {C} \sum\nolimits_{t}\sum\nolimits_{i}(\xi_{i}^{t}+{\xi_i^t}^*) +\frac{1}{2}\operatorname{tr} \mathbf{L L}^{\top}\\
&+ \sum\nolimits_{t}\sum\nolimits_{i}(\alpha_i^t(y_{i}^{t}-\left(\mathbf{L}\mathbf{u}_t\right)^{\top} \phi(\mathbf{x}_{i}^{t})-b_{t}- \xi_{i}^{t}-\varepsilon)- \gamma_i^t \xi_{i}^{t})\nonumber\\
&+\sum\nolimits_{t}\sum\nolimits_{i}({\alpha_i^t}^* (\left(\mathbf{L}\mathbf{u}_t\right)^{\top} \phi(\mathbf{x}_{i}^{t})+b_{t}-y_{i}^{t}-{\xi_i^t}^*-\varepsilon )-{\gamma_i^t}^*{\xi_i^t}^*)\nonumber
\end{eqnarray}
where $\alpha_i^t,{\alpha_i^t}^*, \gamma_i^t, {\gamma_i^t}^* \in \mathbb{R}$ are Lagrangian multipliers.
 
{Then the stationary point conditions are obtained as}
   \[\arraycolsep=1.4pt\def\arraystretch{2.2}
\begin{array}{rll}
\dfrac{\partial \mathcal{L}}{\partial \mathbf{L}}=0 &\Longrightarrow& \mathbf{L}=\sum\nolimits_{t} \sum\nolimits_{i} (\alpha_i^t-{\alpha_i^t}^*) \phi(\mathbf{x}_{i}^{t}) \mathbf{u}_{t}^{\top}, \\
\dfrac{\partial \mathcal{L}}{\partial {b}_t}=0 &\Longrightarrow &   \sum_{i} \alpha_i^t-{\alpha_i^t}^* =0,  \nonumber \\
\dfrac{\partial \mathcal{L}}{\partial \xi_i^t}=0 &\Longrightarrow &   C= \alpha_i^t+\gamma_i^t,  \nonumber  \\
\dfrac{\partial \mathcal{L}}{\partial {\xi_i^t}^*}=0 &\Longrightarrow &  C={\alpha_i^t}^*+{\gamma_i^t}^*,  \nonumber  
\end{array}\] 
By eliminating $\mathbf{L}$ and letting $\lambda_i^t=\alpha_i^t-{\alpha_i^t}^*$, the optimization problem in the dual can be expressed as
\begin{eqnarray}\label{eq: reg-l-sol}
{\max_{\lambda_i^t}} &{-}\frac{1}{2} \sum_{t}  \sum_{q}\sum_{i}\sum_{p} \lambda_i^t \lambda_p^q  \langle\mathbf{u}_t,\mathbf{u}_q\rangle k(\mathbf{x}_i^t,\mathbf{x}_p^q)
{+}\sum_{t}\sum_{i} \lambda_i^t y_i^t+ \lvert \lambda_i^t \lvert \varepsilon \\
\text { s.t. } & \sum_{i} \lambda_i^t =0;  \nonumber\\
& -C\leq  \lambda_i^t  \leq C    \nonumber
\end{eqnarray}
where $\lambda_i^t$ can be {solved}
by this convex QP problem.

\noindent 2) {\bf{{$(\mathbf{U}^{n},b_t,\xi_i^t,{\xi_i^t}^*)$}-subproblem}} With fixed $\mathbf L$, we update $\mathbf{U}^{n},b_t,\xi_i^t,{\xi_i^t}^*$.  The objective function w.r.t.
$\mathbf{u}^{n}_{t_n,:},b_t,\xi_i^t,{\xi_i^t}^*$ {follows from the Lagrangian }
  \begin{eqnarray}\label{eq: reg-u-obj}
&\mathcal{L}(\mathbf{u}^{n}_{t_{n},:}, b_{t}, \xi_{i}^{t},{\xi_i^t}^*)=  {C} \sum_{t\in\mathbb{O}^n_{t_n}} \sum_{i}(\xi_{i}^{t}+{\xi_i^t}^*)  \\
&+ \sum_{t\in\mathbb{O}^n_{t_n}} \sum_{i}\beta_i^t(y_{i}^{t}-{\mathbf{u}_{t_{n},:}^{n} }^{\top}\mathbf{z}_{i}^{t} -b_{t}- \xi_{i}^{t}-\varepsilon)\nonumber\\
&+\sum_{t\in\mathbb{O}^n_{t_n}} \sum_{i}{\beta_i^t}^* ({\mathbf{u}_{t_{n},:}^{n} }^{\top}\mathbf{z}_{i}^{t} +b_{t}-y_{i}^{t}-{\xi_i^t}^*-\varepsilon )\nonumber\\
&+\frac{1}{2}\sum_{n}\lVert \mathbf{u}_{t_{n},:}^{n}\lVert_{2}^2- \sum_{t\in\mathbb{O}^n_{t_n}} \sum_{i}\eta_i^t \xi_{i}^{t}-\sum_{t\in\mathbb{O}^n_{t_n}} \sum_{i}{\eta_i^t}^*{\xi_i^t}^*\nonumber
\end{eqnarray}
where
$\beta_i^t, {\beta_i^t}^*, \eta_i^t,{\eta_i^t}^*$ are the Lagrangian multipliers.

By taking derivatives to the variables, the stationary 
conditions are obtained as 
  \[\arraycolsep=1.4pt\def\arraystretch{2.2}
\begin{array}{rll}
\vspace{0.1cm}
\dfrac{\partial \mathcal{L}}{\partial \mathbf{u}^{n}_{t_{n},:}}=0 &\Longrightarrow& \mathbf{u}^{n}_{t_{n},:}=\sum\limits_{t\in\mathbb{S}_{t_n}} \sum\limits_{i=1}^{m_{t}} (\beta_{i}^{t}-{\beta_i^t}^*) \mathbf{z}_{i}^{t} \nonumber\\ 
\dfrac{\partial \mathcal{L}}{\partial  {b_t}}=0 &\Longrightarrow & \sum_{i=1}^{m_t}\beta_{i}^{t}-{\beta_i^t}^* =0  \nonumber\\
\dfrac{\partial \mathcal{L}}{\partial   \xi_i^t}=0 &\Longrightarrow & C= \beta_i^t+\eta_i^t \nonumber\\ 
\dfrac{\partial \mathcal{L}}{\partial  {\xi_i^t}^*}=0 &\Longrightarrow&  C={\beta_i^t}^*+{\eta_i^t}^*\nonumber
\end{array}\]
By eliminating $\mathbf{u}_{t_n,:}^n$ and letting ${\gamma_i^t}=\beta_i^t-{\beta_i^t}^*$, the
{optimization problem can then be given} by
\begin{eqnarray}\label{eq: reg-u-sol}
{\max_{\gamma_i^t}} &{-}\frac{1}{2} \sum_{t=1}^T  \sum_{q=1}^T\sum_{i=1}^{m_t}\sum_{p=1}^{m_q} \gamma_i^t \gamma_p^q  \langle\mathbf{z}_i^t,\mathbf{z}_p^q\rangle  
{+}\sum_{t=1}^T\sum_{i=1}^{m_t} \gamma_i^t y_i^t+ \lvert \gamma_i^t \lvert \varepsilon \\
\text { s.t. } & \sum_{i=1}^{m_t} \gamma_i^t =0;  \nonumber\\
& -C\leq  \gamma_i^t  \leq C, t\in\mathbb{O}^n_{t_n}   \nonumber
\end{eqnarray}
where $\gamma_i^t$ can be solved by
this convex QP problem. A more detailed derivation process can refer to Appendix section \ref{appendix 1-reg}. The updating procedures are summarized in Algorithm \ref{algorithm:tSVR-MTL}.
\begin{algorithm}
	\caption{tSVR-MTL}
	\label{algorithm:tSVR-MTL}
	\begin{algorithmic}
		\STATE \textbf{Input}: ${y^t_i},x^t_i$ for $t=1,\ldots,T$, $i=1,\cdots,m_t$
		\STATE \textbf{Output}: ${{\lambda_i^t}}, \mathbf{U}^n$ for $n=1,\ldots,N$, and $b_t$ for $t=1,\ldots,T$
		\STATE $\bullet$ Initialize $\mathbf{U}^n$ for $n=1,\ldots,N$
		\WHILE{\textbf{not convergence}}
  		\STATE $\bullet$ Get $\lambda_i^t, {b}_t$ by solving the convex QP problem (\ref{eq: reg-l-sol}) 	
		\FOR{$n=1:N$}
		    \FOR{$t_n=1:T_n$}
		         \STATE $\bullet$ Update $\{\gamma_i^t, {b}_t, t\in\mathbb{O}^n_{t_n},i=1,\cdots,m_t\}$ by solving the convex QP problem (\ref{eq: reg-u-sol})
           \STATE $\bullet$ $\mathbf{u}^{n}_{t_{n},:}=\sum\limits_{t\in\mathbb{S}_{t_n}} \sum\limits_{i=1}^{m_{t}} \theta_{i}^{t} \mathbf{z}_{i}^{t}$
		    \ENDFOR
		\ENDFOR
		\ENDWHILE
	\end{algorithmic}
\end{algorithm}

 \subsection{Decision Function}

The proposed algorithms iterate until reaching the stopping condition, which is set as the relative error of the $k+1$-th update and $k$-th update of factors $\mathbf{U}^{n}$, such that $\sum_n \lVert\mathbf{U}^{n}_{k+1}-\mathbf{U}^{n}_{k}\lVert_{\operatorname{F}}^2/\lVert\mathbf{U}^{n}_{k}\lVert_{\operatorname{F}}^2<10^{-3}$.
 
After optimization,  the  prediction  for any given input $\mathbf x$ of the $t$-th task is obtained either with 
\begin{itemize}
    \item  \textbf{1)  the expression using explicit feature map $\phi(\cdot)$}:
\begin{equation}\label{eq:cla-decision:primal}
f_t(\mathbf{x})=\operatorname{sgn}(\left(\mathbf{L}^*\mathbf{u}_{t}^*\right)^{\top} \phi(\mathbf{x})+b_t^{*}) 
\end{equation}

\item  \textbf{2) the expression  using  kernel function $ k(\cdot, \cdot)$}:  
\begin{equation}\label{eq:cla-decision:dual}
f_t(\mathbf{x})=\operatorname{sgn}(\sum_{p}\sum_{q} \alpha_p^{q*} {y_p^q} k( \mathbf x, \mathbf x_p^q)\langle\mathbf{u}_t^*,\mathbf{u}_q^*\rangle +b_t^{*}).
\end{equation}
\end{itemize}
for classification, where $\alpha_p^{q*}$, $b_t^{*}$ and $\mathbf{u}_t^*$ are derived solution of \eqref{eq:cla-TSVM-MTL:primal}, and
\begin{itemize}
    \item  \textbf{1)  the expression using explicit feature map $\phi(\cdot)$}:
\begin{equation}\label{eq:decision:primal}
f_t(\mathbf{x})=\left(\mathbf{L}^*\mathbf{u}_t^*\right)^{\top} \phi(\mathbf{x})+b_t^{*} 
\end{equation}

\item  \textbf{2) the expression  using  kernel function $ k(\cdot, \cdot)$}:  
\begin{equation}\label{eq:decision:dual}
f_t(\mathbf{x})=\sum_{p}\sum_{q} \lambda_p^{q*} k( \mathbf x, \mathbf x_p^q)\langle\mathbf{u}_t^*,\mathbf{u}_q^*\rangle +b_t^{*}.
\end{equation}
\end{itemize}
for regression, where $\lambda_p^{q*}$, $b_t^{*}$ and $\mathbf{u}_t^*$ are derived solution of \eqref{eq:reg-tsvm-mtl:primal}.
Note that expression 1), {i.e., \eqref{eq:cla-decision:primal}, \eqref{eq:decision:primal},} is the primal representation, while 
expression 2), {i.e., \eqref{eq:cla-decision:dual},\eqref{eq:decision:dual},} is not strictly the dual representation, due to the existence of primal parameters $\mathbf{u}_t^*, \mathbf{u}_q^*$.
This is because the optimization algorithm alternates to update different factors of the coefficient tensor and the resulting Lagrangian dual forms correspond to each subproblem during iterations, not to the original nonconvex problem \eqref{eq:cla-TSVM-MTL:primal} or \eqref{eq:reg-tsvm-mtl:primal}. 

In previous MTL frameworks, the dual expression is normally \begin{equation}
f_t(\mathbf{x})=\operatorname{sgn}(\sum_{p}\sum_{q} \alpha_p^{q*}{y_p^q} k( \mathbf x, \mathbf x_p^q) +b_t^{*}).
\end{equation}
where $\alpha_p^{q*}$ is determined by taking into account the task relatedness. 
However, it can be observed that our derived expression (2) is a weighted expression of the kernel $k(\mathbf{x}, \mathbf{x}_p^q)$, where the weights $\langle\mathbf{u}_t, \mathbf{u}_q\rangle$ are capable of characterizing task-relatedness along different modes. We will further analyze this phenomenon in the experimental section \ref{sec: exp-social}.

\subsection{Efficient training and the LSSVM {variant}}

As shown in Section \ref{sec: tsvm}, the proposed tensorized SVMs for MTL are characterized by a series of iterative convex quadratic programming problems.  {The solution method} of QP problems includes many algorithms, here we use the function "quadprog" in {the Matlab} Optimization toolbox. For a larger problem instance, it could still be time-consuming, {and several efficient approaches are proposed, such as twin SVMs \cite{xie2012multitask} and safe screening rule \cite{wang2014scaling}.  }

To further achieve efficient training, here we employ the learning framework of the Least Squares Support Vector Machine (LSSVM) \cite{suykens1999least} which replaces the inequality constraints with equality constraints {and uses an $\ell_2$ loss function}. This transformation in LSSVMs reduces the original QP problem in classical SVMs to {solving a linear system of equations}, which provides great potentials to speed up the training of SVMs.

By exemplifying  the classification problem, the primal problem of our proposed tensorized methods based on LSSVMs is formulated as
\begin{eqnarray}\label{eq:our:lsstm:primal}
\underset{\mathbf{L}, \mathbf{U}^{n}, b_{t},\xi_i^t}{\min } & \frac{C}{2} \sum_{t}\sum_{i}(\xi_{i}^{t})^{2}+\frac{1}{2}\operatorname{tr} \mathbf{L L}^{\top}+\frac{1}{2}\sum\limits_{n} \operatorname{tr} \mathbf{U}^{n}  {\mathbf{U}^{n}}^{\top} \\ 
\text { s.t. } &
y_{i}^{t}\left( ( \sum_{r} \mathbf{l}_{:,r} {u}_{t,r})^{\top} \phi(\mathbf{x}_{i}^{t})+b_{t}\right) = 1-\xi_{i}^{t} ,\xi_{i}^{t} \geq 0. \nonumber
\end{eqnarray}
Similarly, by formulating {1) $(\mathbf{L}, b_t, \xi_i^t)$-subproblem and 2) $(\mathbf{U}^{n}, b_t, \xi_i^t)$-subproblem with their corresponding stationary conditions,
we can then get the linear systems during the updates concerning
$\mathbf{L}$ and $\mathbf{U}^n$, as follows.}

\noindent 1) The resulting linear system for updating $\mathbf{L}$ is
\begin{equation}    \label{eq: update_L_lp}
    \left[\begin{array}{c|c}\mathbf{0}_{T \times T} & \mathbf{A}^{\top}\mathbf{Y}\\ \hline \mathbf{Y}\mathbf{A} & \mathbf{\mathbf { Q }}+\frac{1}{C} \mathbf{I}_{m\times m}\end{array}\right]\left[\begin{array}{c} \mathbf{b}\\ \boldsymbol{\alpha}\end{array}\right]=\left[\begin{array}{c}\mathbf{0}_T\\ \mathbf{1}_{m}\end{array}\right],
\end{equation}
where 
$\mathbf{Q}\in\mathbb{R}^{m\times m}$ is   computed by 
\begin{equation}\label{eq: lssvc-Q-computation}
    \mathbf{Q}(j,j')=y_i^t y_p^q\left\langle\mathbf{u}_{t}, \mathbf{u}_{q}\right\rangle  k\left(\mathbf{x}_{i}^{t}, \mathbf{x}_{p}^{q}\right).
\end{equation}
And $\mathbf{A}=\text{blockdiag}(\mathbf{1}_{m_1},\cdots,\mathbf{1}_{m_T})\in\mathbb{R}^{m\times T}$, $\mathbf{Y}=\text{diag}(y_1^1,y_2^1,\ldots,y_{m_T}^T)^{\top}\in\mathbb{R}^{m\times m}$, $\mathbf{b}=[b_1,b_2,\ldots,b_T]\in\mathbb{R}^{T}$, 
 and
 ${\boldsymbol{\alpha}}=[\alpha_1^1,\alpha_2^1,\ldots,\alpha_{m_T}^T]^{\top}\in\mathbb{R}^{m}$ denote 
 the outputs, bias and dual variables, $j=\sum_{r=1}^{t-1} m_r +i, j'=\sum_{r=1}^{q-1} m_r +p, i=1,\cdots,m_t, p=1,\cdots,m_q$ with  $i, p$ indexing the  samples in the involved tasks $t$ and $q$, respectively.
 
\noindent 2) The resulting linear system for updating $\mathbf{u}^{n}_{t_n,:}$ is
\begin{equation}  \label{eq: update_U_lp}
    \left[\begin{array}{c|c}
    \mathbf{0}_{\lvert \mathbb{O}_{t_n}\lvert \times \lvert \mathbb{O}_{t_n}\lvert} & \mathbf{A}_{t_n}^{\top}\mathbf{Y}_{t_n}\\ 
      \hline \mathbf{Y}_{t_n}\mathbf{A}_{t_n} & \mathbf{\mathbf { Q_{t_n} }}+\frac{1}{C} \mathbf{I}_{M_{t_n}}\end{array}\right]\left[\begin{array}{c}\mathbf{b}_{t_n}\\
 \boldsymbol{\lambda}_{t_n}\end{array}\right]=\left[\begin{array}{c}\mathbf{0}_{\lvert \mathbb{O}_{t_n}\lvert}\\ \mathbf{1}_{M_{t_n}}\end{array}\right],  
\end{equation}
where $\mathbf{A}_{t_n}=\text{blockdiag}(\mathbf{1}_{m_t})\in\mathbb{R}^{M_{t_n}\times \lvert \mathbb{O}_{t_n}\lvert}$ with $t\in\mathbb{O}_{t_n}$, and ${\mathbf{Y}}_{t_n}$,  ${\boldsymbol{\lambda}}_{t_n}, \mathbf{b}_{t_n} \in \mathbb R^{M_{t_n}}$ are matrices or vectors collecting  $y_i^t$, ${\lambda}_i^t$, and $b_i^t$ involved in the equality constraints, respectively.
Here, the  matrix 
$\mathbf{Q}_{t_n}\in\mathbb{R}^{M_{t_n}\times M_{t_n}}$ is computed by $\mathbf{Q}_{t_n}(j,j')=y_i^ty_p^q\left\langle\mathbf{z}^{t}_i, \mathbf{z}_{p}^q\right\rangle $, 
where $ t, q\in\mathbb{O}_{t_n}, i=1, \ldots, m_{t},p=1,\cdots,m_q$. The detailed derivation process and regression case are illustrated in Appendix Section \ref{appendix 2}.

For this tensorized LSSVMs for MTL, a significant portion of the computational complexity arises from the computation of $\mathbf{Q}$ in \eqref{eq: lssvc-Q-computation} and solving the linear system problem \eqref{eq: update_L_lp} in each iteration. The complexity of solving this linear system is typically $O(m^3)$, where $m$ represents the sample size.

To address this computational challenge, a unified linear system with a positive definite coefficient matrix can be formulated. All the subproblems in \eqref{eq: update_L_lp} and \eqref{eq: update_U_lp} can be represented by the following unified system
\begin{equation}\label{eq: unfied-ls}
    \left[\begin{array}{c|c}
    \mathbf{0}  &  \mathbf{V}^{\top} \\ 
      \hline \mathbf{V} & \mathbf{H}\end{array}\right]\left[\begin{array}{c}\mathbf{x}_1\\
     \mathbf{x}_2\end{array}\right]=\left[\begin{array}{c} \mathbf{d}_1\\ \mathbf{d}_2\end{array}\right],
\end{equation}
where the matrix in \eqref{eq: unfied-ls} is not positive definite. Instead of solving these linear systems where the coefficient matrix is not positive definite directly, we {follow the technique from \cite{johan2002least,suykens1999least} and} formulate these linear systems into one unified linear system with a positive definite coefficient matrix, such that
\begin{equation} 
    \left[\begin{array}{c|c}
    \mathbf{s}  &  \mathbf{0}\\ 
      \hline \mathbf{0} & \mathbf{H}\end{array}\right]\left[\begin{array}{c}\mathbf{x}_1\\
\mathbf{x}_2+\mathbf{H}^{-1}\mathbf{V}\mathbf{x}_1\end{array}\right]=\left[\begin{array}{c} -\mathbf{d}_1+\mathbf{V}^{\top}\mathbf{H}^{-1}\mathbf{d}_2\\ \mathbf{d}_2\end{array}\right],
\end{equation}
where $\mathbf{s}=\mathbf{V}^{\top}\mathbf{H}^{-1}\mathbf{V} >0$.
In this way, fast and efficient methods for linear systems with positive definite matrices can be applied for acceleration. For example, by utilizing Cholesky decomposition, the complexity of solving the linear systems can be reduced to approximately $O(\frac{1}{3}m^3)$.

\section{Experiments}
 \label{sec: experiment}
In this Section, experiments on simulated data and real-world data are conducted {for}
a comprehensive understanding of the effectiveness of the proposed methods, including tSVC-MTL, tSVR-MTL, tLSSVC-MTL and tLSSVR-MTL. For regression, 
the commonly used metrics like root mean squared error (RMSE), $Q^2$ and the correlation of the prediction and ground truth are used to evaluate the prediction accuracy, where $Q^2$ is defined by $Q^2=1-\lVert \mathbf{y}-\hat{\mathbf{y}}\lVert_{\text{F}}^2/\lVert \mathbf{y}\lVert_{\text{F}}^2 $. 
{For classification,} metrics like Accuracy, Precision, Recall, and F1 score are used.

\subsection{Simulated data}
 \textbf{Data generation}
The simulated data {are attained as follows}:
\begin{itemize}
    \item generate the coefficient tensor through the CP form $\mathbf{B}=[\![\mathbf{L}, \mathbf{U}^{1},\cdots,\mathbf{U}^{N}]\!]\in\mathbb{R}^{d\times T_1\times \cdots \times T_N}$, where each entry of $\mathbf{L}$ and $\mathbf{U}^n$ for $n=1,\cdots,N$ is generated randomly from Gaussian distribution $\mathcal{N}(0,1)$;
    \item generate the predictor $\mathbf{X}^t\in\mathbb{R}^{m_t\times d}$ and random noise $\bm \xi^t\in\mathbb{R}^{m_t}$ randomly from distribution $\mathcal{N}(0,1)$;
    \item task-specific bias $b^t$ is sampled from {Gaussian} distribution $\mathcal{N}(0,1)$
    \item let the 
    response $\mathbf{y}^t=\mathbf{X}^t\cdot \sum_{r=1}^R \mathbf{l}_r \cdot u^1_{t_1,r}\cdot \cdots\cdot u^N_{t_N,r}+\text{repmat}({b}_t,m_t,1)$
    \item given signal-to-noise ratio (SNR) defined as
$\text{SNR}=20\log_{10} \frac{\text{var}({\mathbf{y}^t})}{\text{var}(\sigma {\bm \xi^t})}$, generate the corresponding noisy response by $\mathbf{y}^t=\mathbf{y}^t+\sigma \bm \xi^t$ for regression or $\mathbf{y}^t=\text{sign}(\mathbf{y}^t+\sigma \bm \xi^t)$ for classification.
\end{itemize}
 
{In this experiment}, we set $N=3$, $T_1=3$, $T_2=4$, $T_3=5$; so the total number of tasks here is $T=\prod_{n=1}^N T_n=60$. The number of features for each sample is $d=100$. The number of training samples of each task is $60$, the number of testing samples of each task is $20$. The CP rank $R$ is set as $3$, SNR is $10$dB for regression and $40$dB for classification.

\noindent {\textbf{Tuning parameter selection}}
For the proposed tensor-based methods, there are several {hyperparameters to be determined,} 
including the CP rank $R$ and the parameter  {$C$ balancing the trade-off between data fitting error and model complexity}. For this group of experiments, only linear kernels are used. $K$ is selected from $\{1,2,3,4,5\}$, $C$ is selected from $\{2^{-5},2^{-3},\cdots, 2^{13},2^{15}\}$ through a $5$-fold cross validation
on the training data.

\begin{figure}
\centering
         \includegraphics[width=0.8\textwidth]{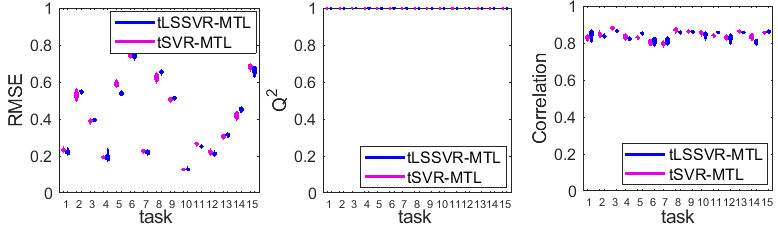} 
    \caption{The effect of random initialization over regression performance.}
    \label{fig:random initial}
\end{figure}

\noindent \textbf{ 
Effect of random initialization} As shown in algorithm \ref{algorithm:tSVM-MTL} and \ref{algorithm:tSVR-MTL}, latent factors $\mathbf{U}^{(n)}$ for $n=1,\cdots,N$ need to be initialized at the beginning and will affect the algorithm performance. In order to analyze the effect of random initialization, we take the regression task as an example and repeat the algorithm $10$ times and report the statistical results in Fig.\ref{fig:random initial}. It gives the statistical results of MSE {versus} different tasks for the proposed methods. Note that here $15$ tasks are selected from all the $60$ tasks for visualization. The experimental results exhibit that random initialization does affect on algorithm performance but seems stable, especially in terms of $Q^2$. 

\begin{figure}
\centering
         \includegraphics[width=0.8\textwidth]{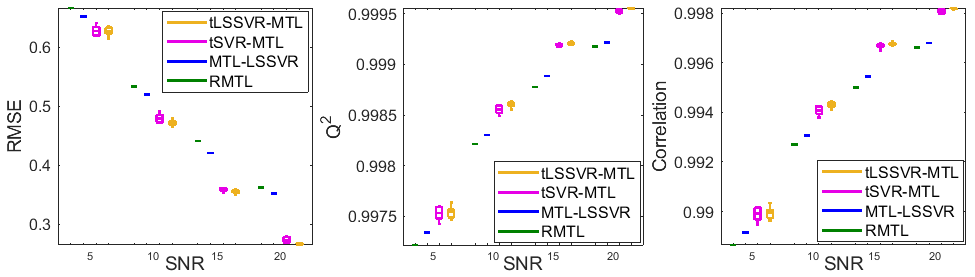} 
    \caption{Performance comparison on simulated data for regression with different SNRs.}
    \label{fig:snr-linear}
\end{figure}

 \begin{figure}
\centering
         \includegraphics[width=0.8\textwidth]{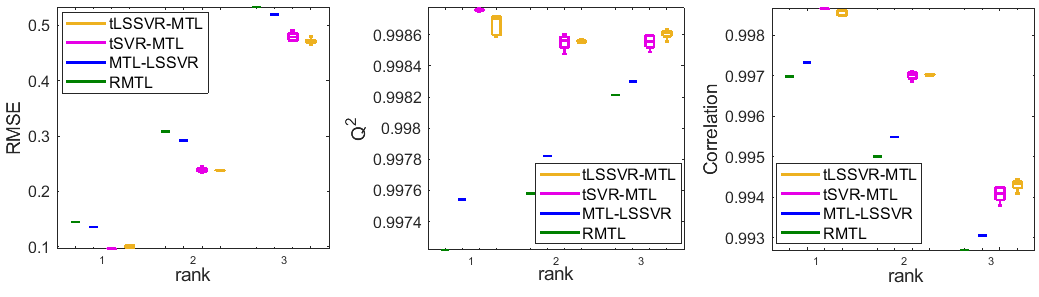} 
    \caption{Performance comparison on simulated data for regression with different {CP} ranks.}
    \label{fig:rank-linear}
\end{figure}
\begin{figure}
\centering
         \includegraphics[width=0.8\textwidth]{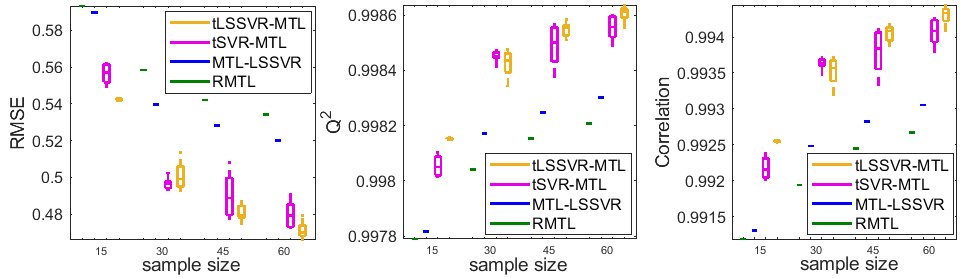} 
    \caption{Performance comparison on simulated data for regression with different sample sizes.}
    \label{fig:sample-linear}
\end{figure}
\noindent  \textbf{Performance improvement w.r.t.
different conditions} This set of experiments aims to analyze the performance improvement of the proposed tensor-based methods compared to other matrix counterparts, {including rMTL \cite{Evgeniou2004regu, Evgeniou2005learn} based on SVMs, and MTL-LSSVC/MTL-LSSVR \cite{xu2014multi} based on LSSVMs}, under various conditions. For both regression and classification tasks, the rank varies in the range of $\{1,2,3\}$, while the sample size ranges from $\{15,30,45,60\}$. For regression, the SNR varies in the range of $\{5,10,15,20\}$, and for classification, it ranges from $\{10,20,30,40\}$.

\begin{figure}
\centering
         \includegraphics[width=0.8\textwidth]{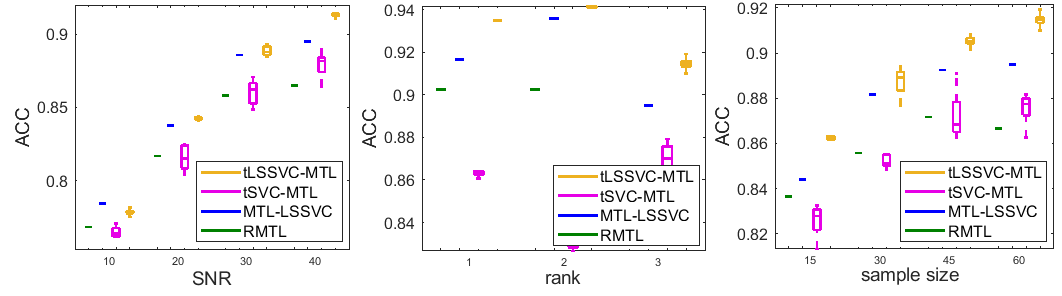} 
    \caption{Performance comparison on simulated data for classification with different conditions.}
    \label{fig:acc-simulated—all}
\end{figure}

Fig. \ref{fig:snr-linear}, Fig. \ref{fig:rank-linear}, and Fig. \ref{fig:sample-linear} present the performance comparison on simulated data for regression, considering different SNRs, ranks, and sample sizes in terms of RMSE, $Q^2$, and correlation, respectively. The experimental results consistently demonstrate that the proposed tensorized methods
give more accurate predictions than their matrix counterparts. Fig. \ref{fig:acc-simulated—all} presents the performance comparison on simulated data for classification, considering different SNRs, ranks, and sample sizes in terms of accuracy (ACC). The experimental results indicate that the proposed tLSSVC-MTL method outperforms MTL-LSSVC in most cases. However, tSVC-MTL requires a higher SNR and sample size to identify appropriate support vectors and construct corresponding hyperplanes.

\subsection{Real Data}

In this group of experiments, we conduct a comprehensive evaluation of the proposed methods by applying them to diverse datasets from social, climate, and medical domains. Our evaluation encompasses both regression and classification tasks, allowing 
to assess the method
effectiveness across different 
data
types. For  all methods based on both SVM and LSSVM frameworks, we explore two kernel options: linear and radial basis function (rbf), where rbf kernel is 
given by 
$k(\mathbf{x},\mathbf{z})=\text{exp}(-p \lVert \mathbf{x}-\mathbf{z}\lVert^2)$, $p>0$ is selected from $\{2^{-15},2^{-13}\cdots, 2^3\}$.

\subsubsection{Social and climate datasets}
\label{sec: exp-social}
In this section, we include four real-world datasets for comparison, consisting of two for regression and two for classification tasks. 
{More details of} these datasets are provided in {the context below and also  Table \ref{tab:real_data_info}.}

   \noindent \textbf{Restaurant \& Consumer.} 
  {This dataset}  \cite{vargas2011effects} contains the rating scores of $138$ consumers to different restaurants based on $45$ various descriptive attributes in terms of 
   food quality, service quality, and overall quality. {Correspondingly, it can be}
   formulated as $138\times 3$ regression tasks. 
   


   \noindent \textbf{CCDS.} The Comprehensive Climate Dataset (CCDS) contains monthly climate records of $17$ variables in North America from 1990 to 2001 \cite{lozano2009spatio}.
   {Here}, $5$ locations are selected and thus
   $5 \times 17$ regression tasks are  for forecasting. 
   {\noindent \textbf{CMU Face Images.}\footnote{\url{http://archive.ics.uci.edu/dataset/124/cmu+face+images}} This dataset comprises face images of 20 persons with varying pose, expression and eyes. In the context of pose estimation (straight, left, right, up) using binary classification, we can create $3\times 3\times 20$  classification tasks by constructing pose pairs using One-vs-One approach.}

\noindent\textbf{Daily and Sports Activities.}\footnote{\url{https://archive.ics.uci.edu/ml/datasets/Daily+and+Sports+Activities}} 
{This} dataset \cite{altun2010comparative} contains records of $19$ daily sports performed by $8$ subjects with a duration of $5$ minutes,
separated into $60$ segments.  
{Here,} $19$ sports are divided into $171$ pairs due to the binary classification setting. Considering the multiple indices for different pairs and 
subjects, 
$18 \times 18 \times 8$ {classification tasks are constructed}. 
    
   \begin{table*}[ht!]
    
    \centering
    \scalebox{0.9}{
   \begin{tabular}{ |c|c|c|c|c|c| }
\hline
Datasets & \# of training samples & \# of training samples for each task  & \# of testing samples & $d$ & $T$\\
\hline
Restaurant \& Consumer&2787& 8 ($\pm$ 3)&696&45&[138,3]\\
CCDS&3468& 41 ($\pm$ 3)&867&5&[5,17]\\
CMU Face Images& 591& 5 ($\pm$ 0) & 1281 & 960 & [3,3,20]\\
Daily and Sports Activities & 3168& 6($\pm$ 0) & 60192 & 1125 & [11,11,8]\\
\hline
\end{tabular}}
    \caption{Dimensional Information for Real Datasets.}
    \label{tab:real_data_info}
\end{table*}



{In this part, we include matrix-based counterparts rMTL \cite{Evgeniou2004regu, Evgeniou2005learn}, MTL-LSSVC/MTL-LSSVR \cite{xu2014multi}, and several tensor-based MTL methods for comparisons, including Convex Multilinear Multitask Learning (MLMTL-C) \cite{romera2013multilinear}, Non-convex Multilinear Multitask Learning (MLMTL-NC) \cite{romera2013multilinear}, and Multilinear Multitask Learning based on the transformed tensor nuclear norm (MLMTL-TTNN) \cite{zhang2023multilinear} using different transformations, such as fast Fourier transform (FFT), discrete cosine transform (DCT), and unitary transform (Data). The prediction results on Restaurant $\&$ Consumer and CCDS datasets by rMTL, MTL-LSSVR with linear and rbf kernels, MLMTL-C, MLMTL-NC, MLMTL-TTNN (FFT, DCT, Data) and proposed tSVR-MTL, tLSSVR-MTL with both linear and rbf kernels are presented in Table \ref{tab:real_exp_1} and \ref{tab:real_exp_2}, respectively.} The best results are 
marked in bold font. The results demonstrate that the proposed method provides more accurate predictions than the corresponding matrix-based methods, particularly achieving a nearly $7\%$ improvement in prediction correlation for the Restaurant \& Consumer Dataset. Additionally, the performance of the proposed methods appears to surpass that of the tensor extensions of a simple linear regression, especially when nonlinear kernels are 
deployed. 
Similarly, in terms of classification, Tables \ref{tab:real_exp_3} and \ref{tab:real_exp_4} report the prediction results by rMTL, MTL-LSSVC with linear and rbf kernels, and proposed tSVC-MTL, tLSSVC-MTL with both linear and rbf kernels on CMU Face Images and Daily and Sports Activities dataset, respectively. The results indicate that the proposed methods are capable of producing more accurate predictions. 

\begin{figure}
\centering
 \includegraphics[width=0.7\textwidth]{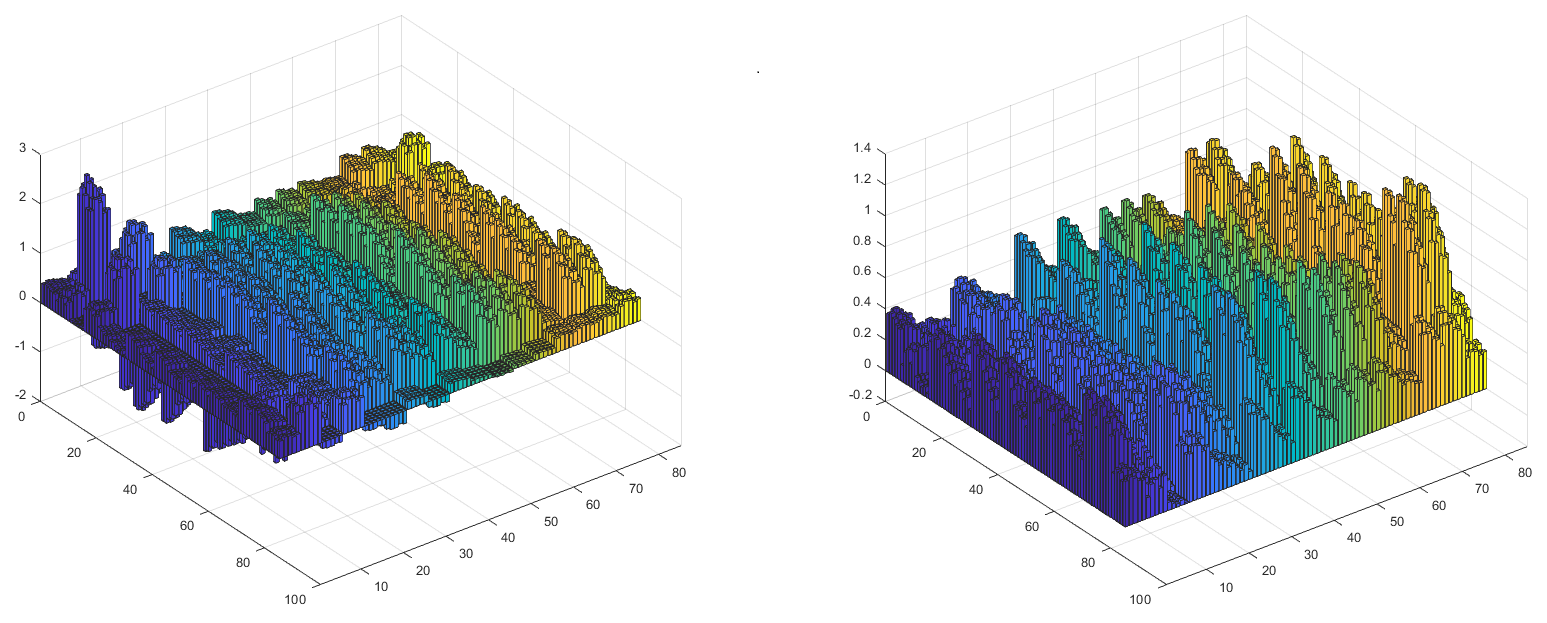} 
    \caption{{Task relatedness characterized by $\langle\mathbf{u}_t, \mathbf{u}_q\rangle, t,q=12,\cdots, T$ (left) and $\langle\mathbf{w}_t, \mathbf{w}_q\rangle, t,q=12,\cdots, T$ (right) on CCDS datasets by tLSSVR-MTL.}}
    \label{fig: task_relatedness}
\end{figure}

Towards more in-depth understandings of the proposed methods, we can investigate the formulated 2) expression, i.e., \eqref{eq:decision:dual}, through exploring the involved factors $\langle\mathbf{u}_t, \mathbf{u}_q\rangle$. { We take the CCDS dataset for an illustrative analysis, and correspondingly
 we plot the results of $\langle\mathbf{u}_t,\mathbf{u}_q\rangle, t,q=12,\cdots, T$ and $\langle\mathbf{w}_t, \mathbf{w}_q\rangle, t,q=12,\cdots, T$ after convergence for comparison in Fig. \ref{fig: task_relatedness} by tLSSVR-MTL.} The Results exhibit the structural consistency of the task-relatedness characterized by $\langle\mathbf{u}_t, \mathbf{u}_q\rangle, t,q=12,\cdots, T$ and $\langle\mathbf{w}_t, \mathbf{w}_q\rangle, t,q=12,\cdots, T$.  We can also observe that the task-relatedness exhibits a block structure, which arises from the natural grouping of tasks. This block structure signifies that certain tasks share stronger connections among themselves compared to other tasks. By leveraging this inherent task grouping, we can effectively exploit the interdependencies and similarities within each block, leading to enhanced performance and {further providing} meaningful insights in multi-task learning {through our tensorized modelling}.

\begin{table}[ht!]
    \centering
    \scalebox{0.95}{{
   \begin{tabular}{ |c|c|c|c|c|c| }
\hline
kernel & Metric & RMSE & $Q^2$ & Correlation \\
\hline
\multirow{8}{*}{linear}&rMTL & 0.71&30.30\%&56.09\%\\
& MTL-LSSVM&0.65&41.83\%&62.54\%\\
&MLMTL-C&0.65& 40.42 \%&61.31\%\\
&MLMTL-NC&0.74 &18.61\%&56.12\%\\
&MLMTL-TTNN(FFT)&0.62 &41.59\%&63.63\%\\
&MLMTL-TTNN(DCT)&\textbf{0.59}&45.90\%&65.18\%\\
&MLMTL-TTNN(Data)&\textbf{0.59} &47.01\%&66.07\%\\
& tSVM-MTL&0.66& 36.64\% & 63.98\%\\
& tLSSTM-MTL &0.61&45.41\%&67.03\%\\
\hline
\multirow{6}{*}{rbf}&rMTL& 0.66&39.30\%&61.31\%\\
&MTL-LSSVM-rbf&0.65&41.90\%&62.55\%\\
& tSVM-MTL&0.65& 38.80\%& 65.50\%\\
&tLSSTM-MTL&\textbf{0.59}&\textbf{49.13}\%&\textbf{69.54}\%\\
\hline
\end{tabular}}}
    \caption{Performance comparison on Restaurant \& Consumer Dataset.}
    \label{tab:real_exp_1}
\end{table}

\begin{table}[ht!]
    \centering
    \scalebox{0.95}{{
   \begin{tabular}{ |c|c|c|c|c|c| }
\hline
kernel &Metric & RMSE & $Q^2$ & Correlation \\
\hline
\multirow{8}{*}{linear}&rMTL & 0.80&27.362\%&54.17\%\\
&MTL-LSSVM & 0.79&29.71\%&55.50\%\\
&MLMTL-C& 0.76&34.56\% &58.79\%\\
&MLMTL-NC&0.83&24.04\%&50.02\%\\
&MLMTL-TTNN(FFT)&0.78&30.75\%&56.86\%\\
&MLMTL-TTNN(DCT)&0.77 &33.92\%&58.70\%\\
&MLMTL-TTNN(Data)&0.77 &33.41\%&58.36\%\\
&tSVM-MTL&0.79&29.69\%&56.50\%\\
&tLSSVM-MTL&0.78&32.64\%&58.03\%\\
\hline
\multirow{6}{*}{rbf}&rMTL& 0.69 & 49.36\% & 70.29\% \\
&MTL-LSSVM& {0.70}&{46.70}\%&{68.36}\%\\
&tSVM-MTL&0.66&51.89\%&72.07\%\\
&tLSSVM-MTL&\textbf{0.65}&\textbf{54.50}\%&\textbf{74.49}\%\\
\hline
\end{tabular}}}
    \caption{Performance comparison on CCDS Dataset.}
    \label{tab:real_exp_2}
\end{table}

\begin{table}[ht!] 
    \centering
    \scalebox{0.95}{{
      \begin{tabular}{ |c|c|c|c|c|c|c| }
\hline
kernel & Metric  & ACC & Precision & Recall & F1 score\\
\hline
\multirow{6}{*}{linear}& rMTL &  90.40\% & 91.59\% & 88.85\% & 90.20\%\\
&MTL-LSSVM  &92.12\% & 92.27\% & 91.84\% & 92.05\% \\
&tSVM-MTL& 93.34\% &93.57\% &93.00\% & 93.28\% \\
&tLSSVM-MTL&\textbf{96.04\%}& \textbf{96.58\% }& \textbf{95.42\% }& \textbf{95.99\%} \\
\hline 
\multirow{6}{*}{rbf}& rMTL  & 58.31\% & 57.54\% & 61.70\% & 59.55\%\\
&MTL-LSSVM  & 68.23\%& 67.58\% & 69.39\% & 68.47\% \\
&tSVM-MTL & 60.90\% & 60.62\% & 61.07\% & 60.82\%\\
&tLSSVM-MTL& 75.43\% & 75.54\%& 74.91\%& 75.18\%  \\
\hline
\end{tabular}}}
    \caption{Performance comparison on CMU face images.}
    \label{tab:real_exp_3}
\end{table}
\begin{table}[ht!]
    
    \centering
    \scalebox{0.95}{
      \begin{tabular}{ |c|c|c|c|c|c|c|c| }
\hline
kernel & Metric &ACC & Precision & Recall & F1 score\\
\hline
\multirow{6}{*}{linear}& rMTL & 85.69\%  & 81.39\% & 92.55\% & 86.61\% \\
&MTL-LSSVM & 85.49\% & 81.32\% & 92.17\% & 86.40\%\\
&tSVM-MTL & 89.10\% & 86.58\% & 92.57\% & 89.48\% \\
&tLSSVM-MTL&91.65\% & 89.62\% & 94.21\% & 91.86\%\\
\hline 
\multirow{6}{*}{rbf}& rMTL & 73.42\% & 74.07\% & 72.13\% & 73.09\% \\
&MTL-LSSVM  & 92.52\% & 93.30\% & 91.62\% & 92.45\% \\

&tSVM-MTL& 94.83\% & 95.00\% & 94.64\% & 94.82\%\\
&tLSSVM-MTL & \textbf{96.40\%}& \textbf{95.98\%} &\textbf{96.86\%}&\textbf{96.42\%}\\
\hline
\end{tabular}}
    \caption{Performance comparison on Daily Activities Dataset.}
    \label{tab:real_exp_4}
\end{table}


\subsubsection{{Application to} disease progression prediction}

Alzheimer's disease (AD) is {one of the most common types of dementia and has long been} a severe neurodegenerative disorder. The identification of biomarkers for tracking the progression of Alzheimer's disease has gained increasing attention in recent years. 
The Mini-Mental State Examination (MMSE) and Alzheimer's Disease Assessment Scale cognitive subscale (ADAS-Cog) are important clinical diagnostic criterions for suspected AD. In this Section, we {apply our proposed methods to} predict clinical parameters related to the progression of AD using patients' basic information such as age, education level, gender, and features extracted from {real-world} MRI images.

Specifically, the data used in this set of experiments are derived from the Alzheimer's Disease Neuroimaging Initiative (ADNI) dataset. Firstly, we obtained diffusion tensor images from different samples and performed a series of preprocessing steps using the FSL software library, including data format conversion, gradient direction extraction, b-value extraction, eddy current correction, motion correction, skull stripping, brain extraction, and so on.
By fitting diffusion tensors to preprocessed images at each pixel, the eigenvalues ($\lambda_1$, $\lambda_2$, $\lambda_3$) of the diffusion tensors are obtained, along with the calculation of fractional anisotropy (FA) and mean diffusivity (MD). After the computation, the obtained eigenvalue tensors, FA, and MD tensors are registered to standard brain regions using the FSL software library. In this experiment, we used the Automated Anatomical Labeling (AAL) parcellation template to divide the obtained feature maps into 90 brain regions and calculate their mean and variance. Considering the three different eigenvalue tensors, FA, and MD, we obtained a total of $90 \times 5 \times2$ features. Additionally, incorporating sample-specific attributes such as age, gender, and years of education, we extracted a total of 903 features for each sample.

In this experiment,
we evaluate the patient's disease progression using four different parameters: ADAS11, ADAS13, MMSE, and DSS (disease severity score). ADAS11 and ADAS13 are both cognitive subscales of the Alzheimer's Disease Assessment Scale, while DSS represents the severity score of the disease, indicating different stages of Alzheimer's disease progression.  Different time points are also considered, including bl (baseline), m06 (6 months after initial visit), and m12 (12 months after initial visit), 
{which constitute $3 \times 4$ tasks in MTL.} 
 \begin{table}
 \centering	 		
 	\scalebox{1}{
 		\begin{tabular}{ |c|c|c|c|c|c| }
   			\hline
 			kernel & Metric & RMSE & $Q^2$ & Correlation   \\
 			\hline
 		\multirow{6}{*}{linear}& rMTL&5.13& 92.31\% & 89.32\% \\
   &MTL-LSSVM & 4.95&92.81\%&90.06\% \\
  & MLMTL-C& 17.95&0.12\% &21.54\% \\
 		&	MLMTL-NC&17.92&0.32\%&10.71\% \\
   & tSVM-MTL&2.70& 97.85\% & 97.13\%\\
 &  tLSSVM-MTL&2.63&97.97\%&97.29\% \\
 \hline
 		\multirow{4}{*}{rbf}&rMTL& 5.89	& 89.86\% & 85.60\%\\
   &MTL-LSSVM& {5.01}&{92.71}\%&{89.95}\% \\
	   & tSVM-MTL& 2.52& 98.13\% &97.51\%\\
 		&	tLSSVM-MTL&\textbf{2.26}&\textbf{98.49}\%&\textbf{98.00}\%  \\
 			\hline
 		\end{tabular}}
   \caption{Performance comparison on ADNI Dataset.}
 	 		\label{tab:real_exp_adni}
 	\end{table}
  
Separating the data
into training set and test set by $4:1$, Table \ref{tab:real_exp_adni} presents the overall prediction performance for all tasks. To further analyze the predictive performance of different algorithms for each task, Table \ref{tab:real_exp_adni_st} represents the RMSE results
achieved by different 
methods,
for ADAS11, ADAS13, MMSE, and DDS scores at different time points. The results demonstrate the effectiveness of the proposed tensorized methods in disease progression prediction. When considering the correlation of prediction results, the proposed method tLSSVR-MTL achieves an $7.94\%$ improvement compared to MTL-LSSVR. Moreover, in comparison to the tensor-based parameterized multitask learning model MLMTL-C and MLMTL-NC, our proposed non-parametric model exhibits substantial improvement. {Note the results by MLMTL-TTNN are not reported since it fails to learn anything and enforces all the weights to zeros.} 
  \begin{table}
 		\centering 				
 		\scalebox{0.88}{
 			\begin{tabular}{ |c|c|c|c|c|c|c|c| }
 				\hline
 				\multirow{2}{*}{kernel}&\multirow{2}{*}{Method}&	\multicolumn{3}{c|}{ ADAS11}&	\multicolumn{3}{c|}{ ADAS13}\\
 				\cline{3-8}
 				&& bl & m06 & m12  & bl & m06 & m12  \\
 				\hline
 			\multirow{6}{*}{linear}&rMTL&3.72& 5.02& 7.14& 6.38& 6.19&8.44\\
    &MTL-LSSVM & 3.35 & 4.33  &6.47  & 5.93& 5.72& 7.89\\
    &MLMTL-C& 10.54&13.40&15.40& 17.60&18.58 &19.94\\
 			&	MLMTL-NC&10.48&13.26&15.18&17.51&18.50&19.81\\
     & tSVM-MTL& 3.00& 2.91&4.01& 3.63& 4.00& 3.67\\
 				&tLSSVM-MTL&2.87&2.62&3.99&3.53&3.79&3.69\\
    \hline
     			\multirow{4}{*}{rbf}&rMTL&3.41& 4.82& 7.14& 6.99& 7.25& 10.27\\
 				&MTL-LSSVM & 3.69 & 5.06  &6.95& 6.43& 6.06& 8.62\\
 				&tSVM-MTL& 2.39& 2.66& 3.70& 3.31& 3.67& 3.85\\
 				&tLSSVM-MTL&\textbf{2.19}&\textbf{2.44} &\textbf{2.99} &\textbf{3.21}&\textbf{3.37} &\textbf{2.97}  \\
 				\hline
 			\end{tabular}
    }
    
 			\scalebox{0.91}{
 				\begin{tabular}{ |c|c|c|c|c|c|c|c| }
 					\hline
 					\multirow{2}{*}{kernel}&\multirow{2}{*}{Method}&	\multicolumn{3}{c|}{ MMSE}&	\multicolumn{3}{c|}{ DDS}\\
 					\cline{3-8}
 					&& bl & m06 & m12  & bl & m06 & m12  \\
 					\hline
       			\multirow{6}{*}{linear}&rMTL&4.17& 5.23& 4.50& 1.67& 1.71& 2.32\\
 					&MTL-LSSVM &4.30& 5.62 &5.08 & 1.90 &2.30 & 2.62\\
      &MLMTL-C& 27.71&27.25&27.25& 3.25&3.25 &3.26\\
 					&MLMTL-NC&27.80&27.32&27.33&3.10&3.22&3.23\\
      &tSVM-MTL& 1.58 & 1.76& 1.84 &\textbf{0.94} & 0.88&\textbf{0.80}\\
 					&tLSSVM-MTL&1.59&1.77&1.82&{0.97}&\textbf{0.87}&{0.81}\\
      \hline
       			\multirow{4}{*}{rbf}&rMTL&4.71& 7.06& 5.98& 1.84& 2.31& 2.66\\
 					&MTL-LSSVM  & 3.62& 4.78 &4.24 & 1.37&1.56 & 1.83 \\
 					&tSVM-MTL&\textbf{ 1.54}& \textbf{1.59}& \textbf{1.74}& 0.96& 0.90& 0.83\\
 					&tLSSVM-MTL &{1.56}& {1.64} & {1.76} & {1.01} &{0.91}&{0.86}\\ 
 					\hline
 				\end{tabular}}
     \caption{Performance comparison on ADNI Dataset for each task in terms of RMSE.}
 				\label{tab:real_exp_adni_st}
 			\end{table}

 \begin{table}[ht!]
    
    \centering
    \scalebox{1}{
      \begin{tabular}{ |c|c|c|c|c|c|c| }
\hline
kernel & Metric  & ACC & Precision & Recall & F1 score\\
\hline
\multirow{6}{*}{linear}& rMTL & 89.69\% & 88.91\% & 90.45\%& 89.67\% \\
&MTL-LSSVM  & 94.22\% & 93.67\% & 94.72\% & 94.19\% \\

&tSVM-MTL& 91.55\% & 90.80\% & 92.29\% & 91.53\% \\
&tLSSVM-MTL &97.63\% & 97.13\% & 98.10\% & 97.61\%\\
\hline 
\multirow{6}{*}{rbf}& rMTL &  91.80\% & 91.26\% & 92.28\% & 91.76\% \\
&MTL-LSSVM &95.42\% & 94.61\% & 96.24\% & 95.42\% \\
&tSVM-MTL & 93.45\% & 92.49\% & 94.45\% & 93.46\% \\
&tLSSVM-MTL& \textbf{98.17\%} & \textbf{97.85\%} & \textbf{98.48\%} & \textbf{98.16\%}\\
\hline
\end{tabular}}
    \caption{Classification performance comparison on ADNI Dataset
    when sampling ratio is $0.3$.}
    \label{tab:real_exp_adni_cls}
\end{table}

\begin{figure}[h]
\centering
          \subfigure[linear]{
\includegraphics[width=0.35\textwidth]{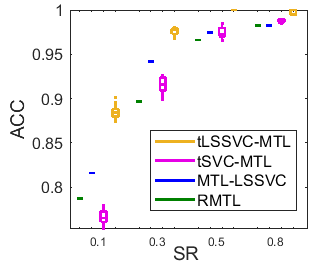} }
     \subfigure[rbf]{
 \includegraphics[width=0.35\textwidth]{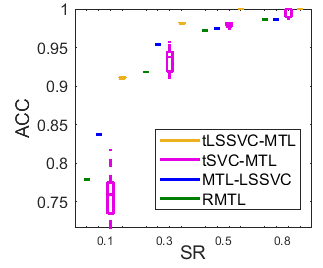} }
    \caption{Classification performance  on ADNI Dataset with different sampling ratios with (a) linear and (b) rbf kernels.}
    \label{fig:acc-adni-linear}
\end{figure}

\begin{figure}[h]
\centering
         \subfigure[ADAS11/ADAS13/DSS]{
\includegraphics[width=0.35\textwidth]{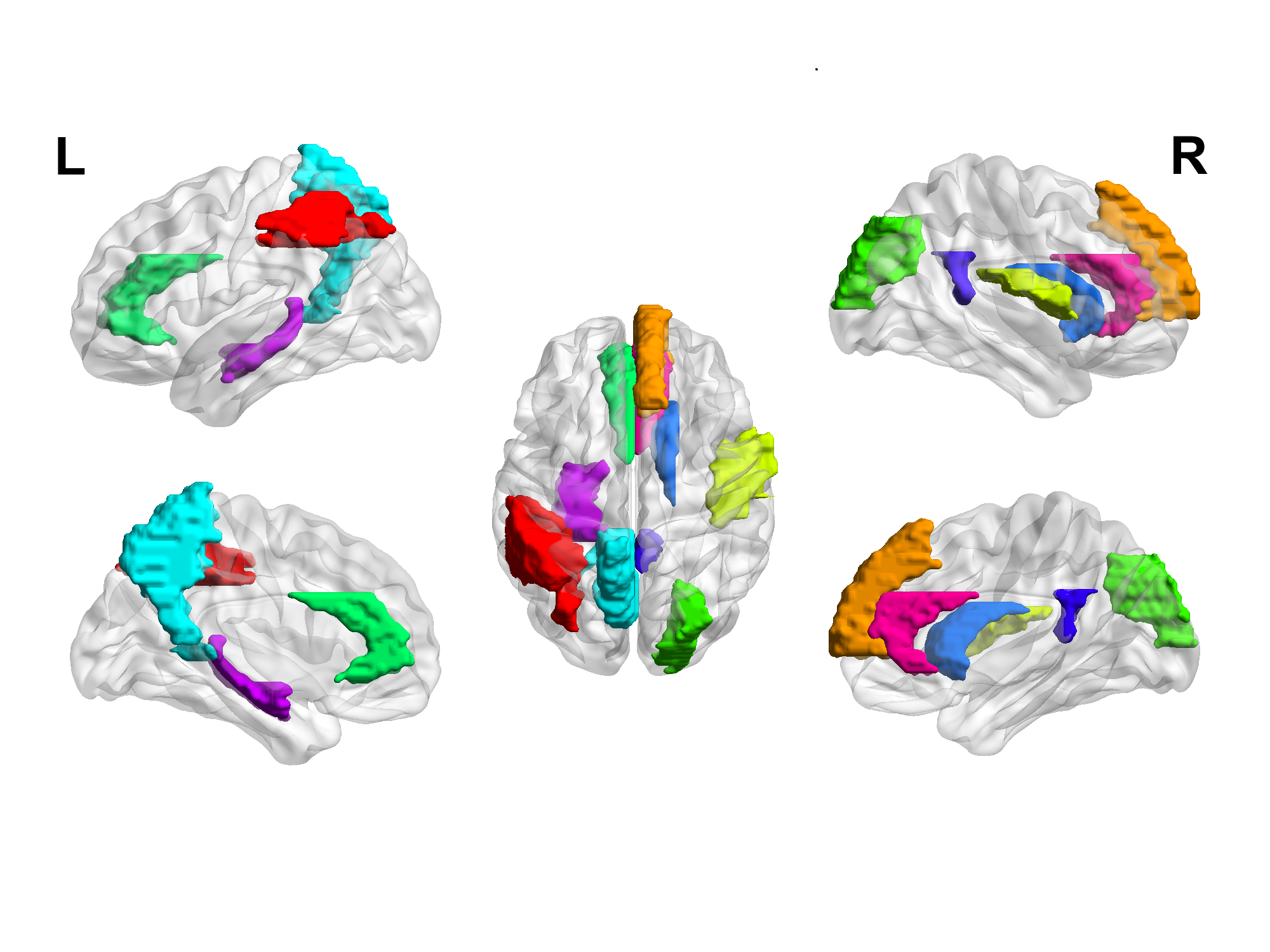}} 
 \subfigure[MMSE]{
\includegraphics[width=0.35\textwidth]{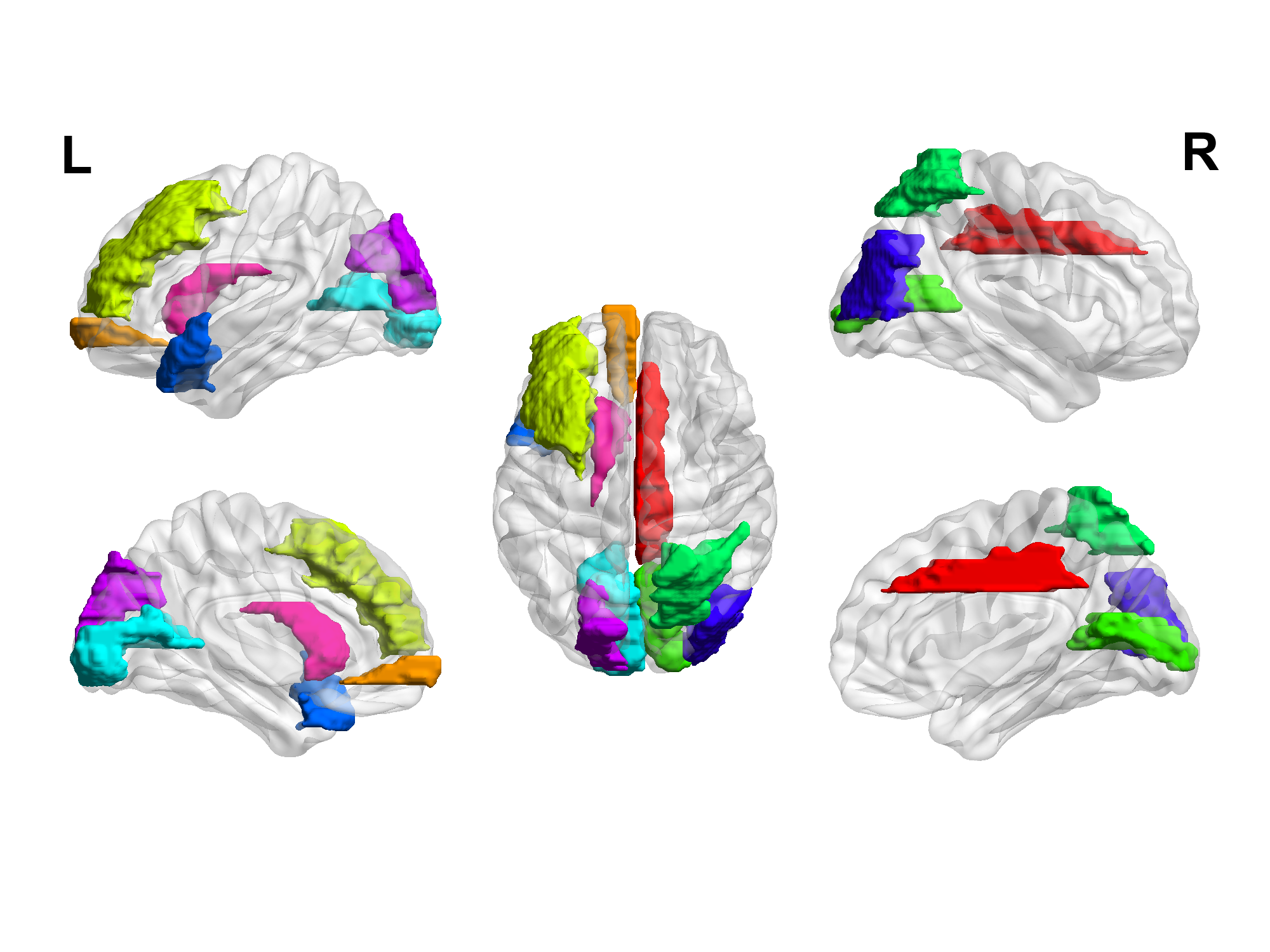} }
    \caption{Top ten important brain regions in the disease progression prediction task using the proposed tLSSVR-MTL algorithm on the ADNI dataset in terms of (a) ADAS11/ADAS13/DSS and (b) MMSE.}
    \label{fig:acc-adni-importance}
\end{figure}

Additionally, considering different time points and different disease stages indicated by DSS, including normal aging/cognitively normal, significant memory concern, early mild cognitive impairment, impairment, and Alzheimer’s disease, we could also construct $4\times 4\times 3$ classification tasks. Table \ref{tab:real_exp_adni_cls} shows the prediction accuracy when the training sample ratio is $0.3$, while Fig. \ref{fig:acc-adni-linear} presents prediction performance by different methods with varied training sample ratios. The results indicate that the proposed tensorized methods achieve better performance compared to their matrix counterparts. However, in cases where the sample ratio is insufficient to identify the appropriate support vectors, the tSVM-MTL algorithm may encounter challenges in constructing an accurate hyperplane for effectively separating different classes of data. Fig. \ref{fig:acc-adni-importance} highlights the top 10
brain regions that influence the disease progression prediction task using the proposed tLSSVR-MTL, where the most important brain regions {affecting} the prediction of ADAS11, ADAS13, and DSS are consistent.

{\subsection{Further Analysis}}
\label{sec: exp-further-analysis}
{\subsubsection{Convergence}}
 
\begin{figure}
\centering
 
\includegraphics[width=0.6\textwidth]{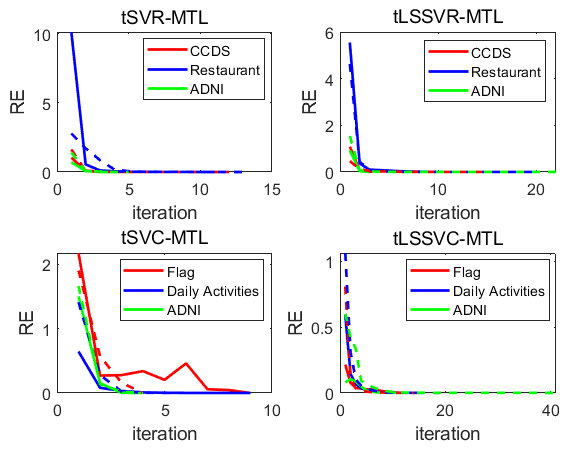} 
    \caption{Convergence performance {on Restaurant \& Consumer, CCDs and ADNI dataset for regression tasks (the top row), and CMU Face Images, Daily Activites and ADNI dataset for classification tasks (the bottom row).}  The solid line corresponds to the linear kernel, while the dotted line corresponds to the rbf kernel. }
    \label{fig:convergence}
\end{figure}

{In Fig. \ref{fig:convergence}, the  relative error between the current and previous steps for factors $\mathbf{U}^n, n=1,\cdots, N$ is plotted over iterations, such that  $\operatorname{RE}=\sum_n \lVert \mathbf{U}_{k+1}^n-\mathbf{U}_k^n\lVert_{\operatorname{F}}^2$ with $k$  the iteration number.
{It presents the convergence curves of proposed methods on Restaurant \& Consumer, CCDs and ADNI dataset for regression tasks, and CMU Face Images, Daily Activites and ADNI dataset for classification tasks.} In this experiment, all algorithms are terminated when the relative error $\operatorname{Re}$ reached $10^{-3}$.} 
The results {empirically} demonstrate that the proposed methods {converge quite fast}
within 
a few iterations. 
 {Moreover, as shown by} 
 the convergence results, it is evident that, in most cases, the stopping condition can be {further} relaxed, allowing for early termination, which can {also help} 
 enhance efficiency.

{\subsubsection{Computational complexity and acceleration}}
\begin{figure}
\centering
 
\includegraphics[width=0.6\textwidth]{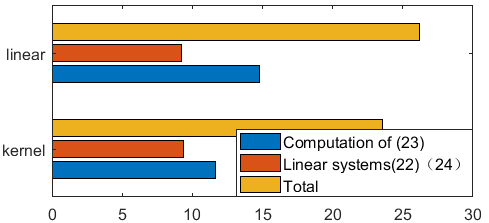} 
    \caption{Complexity analysis over Restaurant \& Consumer dataset. }
    \label{fig:complexity}
\end{figure}

{To analyze the main computational complexity of the proposed tensorized LSSVMs for MTL, we conducted experiments on the Restaurant \& Consumer dataset and carefully recorded the execution time of each step. 
 In Fig. \ref{fig:complexity}, we present a comparison of the execution times for computing the $\mathbf{Q}$ matrix, solving linear systems, and the overall algorithm. This visualization allows us to identify the main source of computational complexity in our method and understand which steps contribute significantly to the overall execution time.} The results confirm that the main complexity of the proposed tensor-based LSSVMs arises from the computation of the $\mathbf{Q}$ matrix and the solution of linear systems, which account for a 
{susbtantial} portion of the training time.

    

In order to speed up the training procedure of proposed tensorized SVMs, we consider two strategies:   1) Cholesky factorization, 2) early stop. The Cholesky factorization is used to tackle linear system $\mathbf{A}\mathbf{X}=\mathbf{B}$ with positive definite matrix $\mathbf{A}$. 
The Cholesky factorization is {firstly} applied as $\mathbf{A}=\mathbf{L}\mathbf{L}^{\top}$ to get the lower triangular factor $\mathbf{L}$ and then $\mathbf{X}$ can be obtained by forward and backward substitution; for early stop, we relax the stopping condition threshold from $10^{-3}$ to $10^{-1}$. 

{Taking} the Restaurant \& Consumer and Daily Sports and Activities datasets {as an illustration}, we present the prediction results along with the
memory and CPU time for different strategies employed in Table \ref{tab:complex_exp_1} and \ref{tab:complex_exp_2}. The results show
that implementing early stopping significantly reduces the training time while maintaining an acceptable level of performance. 
{Furthermore,} the Cholesky factorization can also enhance the efficiency of the algorithm, {and notably does not compromise the} 
prediction performance. 
\begin{table}[ht!]
    
    \centering
    \scalebox{0.95}{
      \begin{tabular}{ |c|c|c|c|c|c|c|c| }
\hline
kernel & \makecell[c]{Cholesky\\ factorization}&\makecell[c]{early\\ stop}      &RMSE& memory &\makecell[c]{CPU\\ Time} \\
\hline
\multirow{6}{*}{linear}& - & -  &0.6155 &3.88e9 & 26.17\\
 & \checkmark & - &0.6155 & 5.76e9& 22.50\\
 & - &  \checkmark  &0.6252& 8.08e8 & 5.42\\
  & \checkmark &  \checkmark  &0.6252& 1.17e9 & 4.34\\
\hline 
\multirow{6}{*}{rbf}& -& - &0.5924 &3.09e9 & 23.55\\
 & \checkmark& -&0.5924 &4.59e9& 17.61\\
 &-& \checkmark &0.5934 & 8.08e8 & 6.05\\
&\checkmark& \checkmark &0.5934 & 1.17e9 & 4.65\\
\hline
\end{tabular}}
    \caption{Performance comparison on Restaurant \& Consumer Dataset.}
    \label{tab:complex_exp_1}
\end{table}
\begin{table}[ht!]
    
    \centering
    \scalebox{0.95}{
      \begin{tabular}{ |c|c|c|c|c|c|c|c| }
\hline
kernel  & \makecell[c]{Cholesky\\ factorization}& \makecell[c]{early\\ stop}   &ACC& memory &\makecell[c]{CPU\\ Time} \\
\hline
\multirow{6}{*}{linear}&- &-  &91.65\% & 3.77e10& 718.10\\
 &\checkmark & -  &91.65\% &5.52e10& 608.89\\
 &-& \checkmark   &87.83\% &7.14e9& 140.60\\
 & \checkmark &  \checkmark &87.83\% &1.00e10& 128.66\\
 \hline 
\multirow{6}{*}{rbf}&- & - &96.40\% & 4.31e10& 800.26\\
 &\checkmark & -  &96.40\% &6.31e10& 742.09\\
 &-& \checkmark   &95.44\% &7.63e9& 275.62\\
 &\checkmark  &  \checkmark &95.44\% & 1.08e10& 194.42\\
 \hline
\end{tabular}}
    \caption{Performance comparison on Daily Activities Dataset.}
    \label{tab:complex_exp_2}
\end{table}

\section{Conclusion}
\label{sec: conclusion}


In this paper, we introduce a novel framework for Multitask Learning (MTL) based on tensorized \ljn{SVMs and LSSVMs}. This framework extends the traditional SVM-based MTL models and enables the consideration of multitasking scenarios involving multiple indices. By utilizing tensor representations for the coefficient tensor, we can effectively exploit similarities and relatedness across different modes {through our structured tensorized factorization. Extensive} 
experimental results
on both simulated data and real-world datasets demonstrate the superiority of our proposed framework over other state-of-the-art methods. {In future work, more in-depth analytical discussions on the convergence behavior of the proposed approach and other acceleration techniques are worthy of further investigation.}

\bibliographystyle{unsrt}  
\bibliography{tsvm}
\appendix

\section{Tensorized SVMs for MTL}
\label{appendix 1}
{With the constructed} tensor representation for {MTL tasks referenced by}  multiple indices, we 
can formulate the corresponding optimization problem {under} the SVM framework for both classification and regression problems.
For clarity, ${t}$ is used 
to denote
the $\overline{t_1,\cdots,t_N}$-th task. 
 \subsection{Classification}
 \label{appendix 1-cla}
First, we consider the tensorized SVMs for classification, i.e., tSVC-MTL.  The primal optimization problem with the deployed low rank CP factorization is formulated as 
\begin{eqnarray}\label{eq:cla-TSVM-MTL:primal}
\underset{\Phi}{ \min } & 
{{C}} \sum_{t} \sum_{i}\xi_{i}^{t}+\frac{1}{2}\operatorname{tr} \mathbf{L L}^{\top}+\frac{1}{2}\sum_{n} \operatorname{tr} \mathbf{U}^{n}  {\mathbf{U}^{n}}^{\top}\\ \nonumber
\text { s.t. } & y_{i}^{t}\left( ( \sum_{r} \mathbf{l}_{:,r}  {u}_{t_{1},r}^{1}  \cdots {u}_{t_{N},r}^{N}  )^{\top} \phi(\mathbf{x}_{i}^{t})+b_{t}\right)\geq 1-\xi_{i}^{t}, \nonumber\\
& \xi_{i}^{t} \geq 0,  t=\overline{t_{1},\cdots,t_{N}}  \nonumber.
\end{eqnarray}
where $\Phi=\{\mathbf{L}, \mathbf{U}^{n}, b_{t}\}$ is the collection of unknown variables, $t=1,\cdots,T$, $i=1,\cdots,m_t$, $r=1,\cdots,R$.  

As shown in 
\eqref{eq:cla-TSVM-MTL:primal}, the product operations between the shared factor $\mathbf{L}$ and the task-specific factors $\{\mathbf{U}^1,\ldots, \mathbf{U}^N\}$ of the coefficient tensor
{lead to a nonconvex problem in optimization.}
In order to solve this problem, we  update each latent factor while the others are fixed based on the alternating minimization technique,
i.e., the block coordinate descent.

\noindent {\bf{ 1) {$(\mathbf{L}, b_t, \xi_i^t)$}-subproblem}}
 The optimization  w.r.t. $\mathbf{L}, b_t,\xi_i^t$ in the primal is formulated as
\begin{eqnarray} 
\underset{\mathbf{L}, b_{t},\xi_i^t}{ \min } & {{C}}\sum_{t=1}^{T} \sum_{i=1}^{m_{t}}\xi_{i}^{t}+\frac{1}{2}\operatorname{tr} \mathbf{L L}^{\top}\\ 
\text { s.t. } & y_{i}^{t}\left( ( \sum_{r} \mathbf{l}_{:,r} {u}_{t,r})^{\top} \phi(\mathbf{x}_{i}^{t})+b_{t}\right) \geq 1-\xi_{i}^{t} ,\xi_{i}^{t} \geq 0 \nonumber
\end{eqnarray}
where $u_{t,r}\triangleq {u}_{t_{1},r}^{1}  \cdots {u}_{t_{N},r}^{N}$ for $t=\overline{t_{1},\ldots,t_{N}}, t_{n}=1,\ldots,T_{n}, n=1, \ldots, N$, $i=1,\ldots, m_t$.

With dual variables $\alpha_i^t$, $\gamma_i^t$, the Lagrangian function is 
\begin{eqnarray}
\mathcal{L}\left(\mathbf{L}, b_{t},\alpha_i^t, \gamma_i^t\right)
&&={{C}} \sum_{t=1}^{T} \sum_{i=1}^{m_{t}}\xi_{i}^{t}+\frac{1}{2}\operatorname{tr} \mathbf{L L}^{\top}-\sum_{t=1}^{T}\sum_{i=1}^{m_{t}} \gamma_i^t \xi_i^t\\
&&- \sum_{t=1}^{T}\sum_{i=1}^{m_{t}}\alpha_{i}^{t}[ y_{i}^{t}(\mathbf{u}_{t}^{\top} \mathbf{L} ^{\top}\phi(\mathbf{x}_{i}^{t})+b_{t})-1+\xi_{i}^{t} ]\nonumber
\end{eqnarray}
where  $\mathbf{u}_{t} \triangleq [u_{t,1}, \ldots, u_{t, R}]^{\top}\in\mathbb{R}^{R}$.

{The KKT conditions yield the stationary points, such that}
 \[\arraycolsep=1.4pt\def\arraystretch{2.2}
\begin{array}{l}
  \dfrac{\partial \mathcal{L}}{\partial \mathbf{L}}=0 \Longrightarrow \mathbf{L}=\sum_{t} \sum_{i} \alpha_{i}^{t} y_{i}^{t} \phi(\mathbf{x}_{i}^{t}) \mathbf{u}_{t}^{\top} \nonumber\\ 
\dfrac{\partial \mathcal{L}}{\partial {b}_t}=0\Longrightarrow \sum_{i=1}^{m_t} \alpha_i^ty_i^t=0 \nonumber  \\
\dfrac{\partial \mathcal{L}}{\partial \xi_i^t}=0 \Longrightarrow   C  ={\alpha_i^t}+  \gamma_i^t.\nonumber   
\end{array}
\]
The dual problem can then be expressed as
\begin{eqnarray}
\max_{\alpha_i^t} &-\frac{1}{2} \sum_{t} \sum_{q}\sum_{i=1}^{m_t}\sum_{p=1}^{m_q} \alpha_i^t\alpha_p^q y_i^t y_p^q \langle\mathbf{u}_t,\mathbf{u}_q\rangle k(x_i^t,x_p^q) \label{eq: cla_update_L}
+\sum_{t=1}^{T} \sum_{i=1}^{m_t} \alpha_i^t \\
\text { s.t. } &  \sum_{i=1}^{m_t} \alpha_i^ty_i^t=0, 0\leq \alpha_i^t \leq C, 
\nonumber
\end{eqnarray}
where $\alpha_i^t$ can be obtained by solving this convex QP problem, $t,q=1,\cdots,T$.  

\noindent {\bf{ 2)  {$(\mathbf{U}^{n}, b_t, \xi_i^t)$}-subproblem}} 
{Acccordingly,} the update of factor $\mathbf{U}^n\in\mathbb{R}^{T_n\times R}$ can be divided into $T_n$ independent subproblems along the first mode.  Specially, the primal problem w.r.t. $\mathbf{U}^{n}_{t_n,:}\in\mathbb{R}^{R}$ is 
\begin{eqnarray}
\underset{ \mathbf{u}^{n}_{t_n,:}, b_{t}, \bm \xi}{ \min } & {{C}} \sum_{t\in\mathbb{O}^n_{t_n}} \sum_{i=1}^{m_{t}}\xi_{i}^{t}+\frac{1}{2}\lVert \mathbf{u}_{t_{n},:}^{n}\lVert_{2}^2\nonumber\\ \nonumber
\text { s.t. } & y_{i}^{t}\left(  {\mathbf{u}_{t_{n},:}^{n} }^{\top}\mathbf{z}_{i}^{t} +b_{t}\right)\geq 1-\xi_{i}^{t}, \xi_{i}^{t} \geq 0, t\in\mathbb{O}^n_{t_n}, \nonumber
\end{eqnarray}
where $\mathbf{z}_{i}^{t}$ is  formulated as $\mathbf{L}^{\top}  {\phi(\mathbf{x}_{i}^{t}) } \odot \mathbf{u}_{t_{1},:}^1\odot\cdots\odot \mathbf{u}_{t_{n-1},:}^{n-1}\odot\mathbf{u}_{t_{n+1},:}^{n+1}\odot\cdots\odot \mathbf{u}_{t_{N},:}^N\in\mathbb{R}^R$,  the index set containing the involved tasks is given by
$\mathbb{O}^n_{t_n}=\{\overline{t_{1}, \ldots, t_{N}}| t_{l}=1,\ldots,T_{l}, l=1, \ldots, N, l \neq n\}$ with cardinality
$|\mathbb{O}^n_{t_n}|=\prod_{l,l\neq n}T_l$, and the indices in constraints satisfy $t\in\mathbb{O}^n_{t_n}, i=1, \ldots, m_t$.  Here the total number of constraints is ${M^n_{t_n}}=\sum_{t\in \mathbb{O}^n_{t_n}} m_t$. 

With dual variables $\beta_i^t$, $\eta_i^t$, the Lagrangian function is 

\begin{eqnarray}
\mathcal{L}\left(\mathbf{u}^{n}_{t_{n},:}, b_{t}, \bm \xi\right)= {\boldsymbol{C}} \sum_{t\in\mathbb{O}^n_{t_n}} \sum_{i=1}^{m_{t}}\xi_{i}^{t}+\frac{1}{2}\lVert \mathbf{u}_{t_{n},:}^{n}\lVert_{2}^2 -\sum_{t\in\mathbb{O}^n_{t_n}} \sum_{i=1}^{m_{t}}\beta_{i}^{t}[y_{i}^{t}\left(  {\mathbf{u}_{t_{n},:}^{n} }^{\top}\mathbf{z}_{i}^{t} +b_{t}\right)- 1+\xi_{i}^{t}] -\sum_{t\in\mathbb{O}^n_{t_n}} \sum_{i=1}^{m_{t}}\eta_i^t \xi_{i}^{t} \nonumber
\end{eqnarray}


Finding the stationary points we get 
 \[\arraycolsep=1.4pt\def\arraystretch{2.2}
\begin{array}{rll}
\dfrac{\partial \mathcal{L}}{\partial \mathbf{u}^{n}_{t_{n},:}}=0 &\Longrightarrow&\mathbf{u}^{n}_{t_{n},:}=\sum_{t\in\mathbb{O}^n_{t_n}}\sum_{i=1}^{m_t} \beta_{i}^{t} y_{i}^{t} \mathbf{z}_{i}^{t} \nonumber\\ 
\dfrac{\partial \mathcal{L}}{\partial b_t}=0 &\Longrightarrow& \sum_{i=1}^{m_t} y_i^t \beta_i^t=0, \nonumber\\
\dfrac{\partial \mathcal{L}}{\partial \xi_i^t}=0  &\Longrightarrow &C  ={\beta_i^t}+  \eta_i^t, \nonumber
\end{array}\]
where $t\in\mathbb{O}_{t_n}^n$. 

The dual problem can be expressed as
\begin{eqnarray}
\max_{\bm\beta^t_{i}} &&-\frac{1}{2} \sum_{t\in\mathbb{O}_{t_n}^n}  \sum_{q\in\mathbb{O}_{t_n}^n}\sum_{i=1}^{m_t}\sum_{p=1}^{m_q} \beta_i^t\beta_p^q y_i^t y_p^q \langle\mathbf{z}_i^t,\mathbf{z}^q_p\rangle\label{eq: cla_update_U}+\sum_{t\in\mathbb{O}_{t_n}^n} \sum_{i=1}^{m_t} \beta_i^t \\
\text { s.t. } && \sum_{i=1}^{m_t} \beta_i^ty_i^t=0;  \nonumber\\
&& 0\leq \beta_i^t \leq C  \text{ for } t\in\mathbb{O}_{t_n}^n, i=1,\cdots, m_t   \nonumber
\end{eqnarray}
where $\beta_i^t,t\in\mathbb{O}_{t_n}^n$ can be obtained by solving this convex QP problem.  

\subsection{Regression}
\label{appendix 1-reg}
 
{For regression, the optimization problem of}
our tensorized SVMs,
i.e., tSVR-MTL, is constructed in the primal as: 
\begin{eqnarray}\label{eq:our:tsvr-mtl:primal}
\underset{\Phi}{\min } &  {C} \sum_{t}\sum_{i}(\xi_{i}^{t}+{\xi_i^t}^*) +\frac{1}{2}\operatorname{tr} \mathbf{L L}^{\top}+\frac{1}{2}\sum_{n} \operatorname{tr} \mathbf{U}^{n}  {\mathbf{U}^{n}}^{\top} 
\quad\\ 
\text { s.t. } &
f(\mathbf{x}_i^t)=( \sum\nolimits_{k} \mathbf{l}_{:,k}  \cdot {u}_{t_{1},k}^{1}  \ldots {u}_{t_{N},k}^{N}  )^{\top} \phi(\mathbf{x}_{i}^{t})+b_{t} \nonumber\\
& y_{i}^{t}-f(\mathbf{x}_i^t)\leq \xi_{i}^{t}+\varepsilon ,\nonumber \\
& f(\mathbf{x}_i^t)-y_{i}^{t}\leq {\xi_{i}^{t}}^*+\varepsilon ,\nonumber\\
& \xi_{i}^{t}, {\xi_i^t}^*\geq 0,\quad t=\overline{t_1,\ldots,t_N}. \nonumber
\end{eqnarray}
where $\Phi=\{\mathbf{L}, \{\mathbf{U}^{n}\}_{n=1}^N, \{b_{t}\}_{t=1}^T\}$ is the collection of unknown variables to be determined.

\noindent 1) {\bf{{$(\mathbf{L},b_t,\xi_i^t,{\xi_i^t}^*)$}-subproblem}} 
The optimization w.r.t. 
$\mathbf{L},b_t,\xi_i^t,{\xi_i^t}^*$ is 
\begin{eqnarray}\label{eq:tsvr-update:l}
\underset{\mathbf{L}, \mathbf{U}^{n}, b_{t},\xi_i^t}{\min } &  {C} \sum\limits_{t=1}^T\sum\limits_{i=1}^{m_t}(\xi_{i}^{t}+{\xi_i^t}^*) +\frac{1}{2}\operatorname{tr} \mathbf{L L}^{\top} \\ 
\text { s.t. } &
f(\mathbf{x}_i^t)=\left(\mathbf{L}\mathbf{u}_t\right)^{\top} \phi(\mathbf{x}_{i}^{t})+b_{t} \nonumber\\
& y_{i}^{t}-f(\mathbf{x}_i^t)\leq \xi_{i}^{t}+\varepsilon ,\nonumber \\
& f(\mathbf{x}_i^t)-y_{i}^{t}\leq {\xi_i^t}^*+\varepsilon ,\nonumber\\
& \xi_{i}^{t},{\xi_i^t}^*\geq 0,\quad t=\overline{t_1,\ldots,t_N}. \nonumber
\end{eqnarray}

With dual variables $\alpha_i^t,{\alpha_i^t}^*, \gamma_i^t, {\gamma_i^t}^* \in \mathbb{R}$, the Lagrangian function is 
\begin{eqnarray}
&&\mathcal{L}({\mathbf{L}, \mathbf{U}^{n}, b_{t},\xi_i^t})=  {C} \sum\limits_{t=1}^T\sum\limits_{i=1}^{m_t}(\xi_{i}^{t}+{\xi_i^t}^*) +\frac{1}{2}\operatorname{tr} \mathbf{L L}^{\top}+ \alpha_i^t(y_{i}^{t}-\left(\mathbf{L}\mathbf{u}_t\right)^{\top} \phi(\mathbf{x}_{i}^{t})-b_{t}- \xi_{i}^{t}-\varepsilon)\nonumber\\
&&\quad+{\alpha_i^t}^* (\left(\mathbf{L}\mathbf{u}_t\right)^{\top} \phi(\mathbf{x}_{i}^{t})+b_{t}-y_{i}^{t}-{\xi_i^t}^*-\varepsilon )- \gamma_i^t \xi_{i}^{t}-{\gamma_i^t}^*{\xi_i^t}^*\nonumber
\end{eqnarray}


{The KKT conditions yield the stationary points, such that}
 \[\arraycolsep=1.4pt\def\arraystretch{2.2}\begin{array}{rll}
\dfrac{\partial \mathcal{L}}{\partial \mathbf{L}}=0 &\Longrightarrow& \mathbf{L}=\sum\limits_{t=1}^T \sum\limits_{i=1}^{m_t} (\alpha_i^t-{\alpha_i^t}^*) \phi(\mathbf{x}_{i}^{t}) \mathbf{u}_{t}^{\top}, \\ 
\dfrac{\partial \mathcal{L}}{\partial {b}_t}=0 &\Longrightarrow &   \sum_{i=1}^{m_t} \alpha_i^t-{\alpha_i^t}^* =0,  \nonumber \\
\dfrac{\partial \mathcal{L}}{\partial \xi_i^t}=0 &\Longrightarrow &   C= \alpha_i^t+\gamma_i^t,  \nonumber  \\
\dfrac{\partial \mathcal{L}}{\partial {\xi_i^t}^*}=0 &\Longrightarrow &  C={\alpha_i^t}^*+{\gamma_i^t}^*,  \nonumber  
\end{array}\]


By eliminating $\mathbf{L}$ and letting $\lambda_i^t=\alpha_i^t-{\alpha_i^t}^*$, the optimization problem in the dual can be expressed as
\begin{eqnarray}\label{eq:tsvr-sol-l}
\max_{\lambda_i^t} &-\frac{1}{2} \sum_{t=1}^T  \sum_{q=1}^T\sum_{i=1}^{m_t}\sum_{p=1}^{m_q} \lambda_i^t \lambda_p^q  \langle\mathbf{u}_t,\mathbf{u}_q\rangle k(\mathbf{x}_i^t,\mathbf{x}_p^q)
+\sum_{t=1}^T\sum_{i=1}^{m_t} \lambda_i^t y_i^t+ \lvert \lambda_i^t \lvert \varepsilon \\
\text { s.t. } & \sum_{i=1}^{m_t} \lambda_i^t =0;  \nonumber\\
& -C\leq  \lambda_i^t  \leq C    \nonumber
\end{eqnarray}
where $\lambda_i^t$ can be {solved}
by 
this convex QP problem. 

\noindent 2) {\bf{{$(\mathbf{U}^{n},b_t,\xi_i^t,{\xi_i^t}^*)$}-subproblem}} With fixed $\mathbf L$, we update $\mathbf{U}^{n},b_t,\xi_i^t$. 
The resulting optimization problem w.r.t.
$\mathbf{u}^{n}_{t_n,:},b_t,\xi_i^t,{\xi_i^t}^*$ is 
\begin{eqnarray}\label{eq:our:tsvr:update-u}
\underset{ \mathbf{u}^{n}_{t_n,:},b_{t},\xi_i^t,{\xi_i^t}^*}{\min } &  {C} \sum\limits_{t\in\mathbb{O}^n_{t_n} }\sum\limits_{i=1}^{m_t}(\xi_{i}^{t}+{\xi_i^t}^*) +\frac{1}{2}\sum\limits_{n=1}^{N}\lVert \mathbf{u}_{t_{n},:}^{n}\lVert_{2}^2 \\ 
\text { s.t. } &
f(\mathbf{x}_i^t)={\mathbf{u}_{t_{n},:}^{n} }^{\top}\mathbf{z}_{i}^{t} +b_{t} \nonumber\\
& y_{i}^{t}-f(\mathbf{x}_i^t)\leq \xi_{i}^{t}+\varepsilon ,\nonumber \\
& f(\mathbf{x}_i^t)-y_{i}^{t}\leq {\xi_{i}^{t}}^*+\varepsilon ,\nonumber\\
& \xi_{i}^{t}, {\xi_i^t}^*\geq 0,\quad t=\overline{t_1,\ldots,t_N}. \nonumber
\end{eqnarray}


By introducing dual variables 
$\beta_i^t, {\beta_i^t}^*, \eta_i^t,{\eta_i^t}^*$, the Lagrangian function is  formulated as
\begin{equation}
\begin{array}{rl}
&\mathcal{L}\left(\mathbf{u}^{n}_{t_{n},:}, b_{t}, \xi_{i}^{t}\right)=  {C} \sum\limits_{t\in\mathbb{O}^n_{t_n}} \sum\limits_{i=1}^{m_t}(\xi_{i}^{t}+{\xi_i^t}^*) +\frac{1}{2}\sum\limits_{n=1}^{N}\lVert \mathbf{u}_{t_{n},:}^{n}\lVert_{2}^2 + \sum\limits_{t\in\mathbb{O}^n_{t_n}} \sum\limits_{i=1}^{m_t}\beta_i^t(y_{i}^{t}-{\mathbf{u}_{t_{n},:}^{n} }^{\top}\mathbf{z}_{i}^{t} -b_{t}- \xi_{i}^{t}-\varepsilon)\nonumber\\
&\quad+\sum\limits_{t\in\mathbb{O}^n_{t_n}} \sum\limits_{i=1}^{m_t}{\beta_i^t}^* ({\mathbf{u}_{t_{n},:}^{n} }^{\top}\mathbf{z}_{i}^{t} +b_{t}-y_{i}^{t}-{\xi_i^t}^*-\varepsilon )- \sum\limits_{t\in\mathbb{O}^n_{t_n}} \sum\limits_{i=1}^{m_t}\eta_i^t \xi_{i}^{t}-\sum\limits_{t\in\mathbb{O}^n_{t_n}} \sum\limits_{i=1}^{m_t}{\eta_i^t}^*{\xi_i^t}^*\nonumber
\end{array}
\end{equation}

By taking derivatives to the variables, the stationary 
conditions are obtained as 
 \[\arraycolsep=1.4pt\def\arraystretch{2.2}
\begin{array}{rll}
\vspace{0.1cm}
\dfrac{\partial \mathcal{L}}{\partial \mathbf{u}^{n}_{t_{n},:}}=0 &\Longrightarrow& \mathbf{u}^{n}_{t_{n},:}=\sum\limits_{t\in\mathbb{O}_{t_n}} \sum\limits_{i=1}^{m_{t}} (\beta_{i}^{t}-{\beta_i^t}^*) \mathbf{z}_{i}^{t} \nonumber\\ 
\dfrac{\partial \mathcal{L}}{\partial  {b_t}}=0 &\Longrightarrow & \sum_{i=1}^{m_t}\beta_{i}^{t}-{\beta_i^t}^* =0  \nonumber\\
\dfrac{\partial \mathcal{L}}{\partial   \xi_i^t}=0 &\Longrightarrow & C= \beta_i^t+\eta_i^t \nonumber\\ 
\dfrac{\partial \mathcal{L}}{\partial  {\xi_i^t}^*}=0 &\Longrightarrow&  C={\beta_i^t}^*+{\eta_i^t}^*\nonumber
\end{array}\]

 

By eliminating $\mathbf{u}_{t_n,:}^n$ and letting $\gamma_i^t=\beta_i^t-{\beta_i^t}^*$, the
{optimization problem can then be given} by
\begin{eqnarray}\label{eq:tsvr-sol-u}
\max_{\gamma_i^t} &-\frac{1}{2} \sum_{t=1}^T  \sum_{q=1}^T\sum_{i=1}^{m_t}\sum_{p=1}^{m_q} \gamma_i^t \gamma_p^q  \langle\mathbf{z}_i^t,\mathbf{z}_p^q\rangle  
+\sum_{t=1}^T\sum_{i=1}^{m_t} \gamma_i^t y_i^t+ \lvert \gamma_i^t \lvert \varepsilon \\
\text { s.t. } & \sum_{i=1}^{m_t} \gamma_i^t =0;  \nonumber\\
& -C\leq  \gamma_i^t  \leq C, t\in\mathbb{O}^n_{t_n}   \nonumber
\end{eqnarray}
where $\gamma_i^t$ can be solved by
this convex QP problem.  

\section{LSSVM Variant}
 \label{appendix 2}
\subsection{Classification}
 {By exemplifying  the classification problem,} the primal problem {of our proposed tensorized methods} based on LSSVMs is formulated as
\begin{eqnarray}\label{eq:our:lsstm:primal}
\underset{\mathbf{L}, \mathbf{U}^{n}, b_{t},\xi_i^t}{\min } & \frac{C}{2} \sum\limits_{t=1}^T\sum\limits_{i=1}^{m_t}(\xi_{i}^{t})^{2}+\frac{1}{2}\operatorname{tr} \mathbf{L L}^{\top}+\frac{1}{2}\sum\limits_{n=1}^{N} \operatorname{tr} \mathbf{U}^{n}  {\mathbf{U}^{n}}^{\top} \\ 
\text { s.t. } &
y_{i}^{t}\left( ( \sum_{r} \mathbf{l}_{:,r} {u}_{t,r})^{\top} \phi(\mathbf{x}_{i}^{t})+b_{t}\right) = 1-\xi_{i}^{t} ,\xi_{i}^{t} \geq 0. \nonumber
\end{eqnarray}
Similarly, we can solve this problem by alternating minimization.



\vspace{0.2cm}
\noindent 1) {\bf{$\mathbf{L},b_t,\xi_i^t$-subproblem.}}
 
The primal problem with respect to $\mathbf{L},b^t,\xi_i^t$ is given by
\begin{eqnarray}\label{eq:update:l}
\underset{\mathbf{L}, b_{t},\xi_i^t}{\min } & \frac{{C}}{2} \sum\nolimits_{t=1}^{T} \sum\nolimits_{i=1}^{m_{t}}\left(\xi_i^t\right)^{2}+\frac{1}{2}\operatorname{tr} (\mathbf {L L}^{\top})\nonumber\\
\text { s.~t. } & 
y_{i}^{t}(( \sum\nolimits_{k=1}^K \left(\mathbf{l}_{:,k}  \cdot {u}_{t,k}\right))^{\top} \phi(\mathbf{x}_{i}^{t})+b_{t})
=1-\xi_i^t, \nonumber
\end{eqnarray}
where $u_{t,k}\triangleq {u}_{t_{1},k}^{1}  \cdots {u}_{t_{N},k}^{N}$ for $t=\overline{t_{1},\ldots,t_{N}}, t_{n}=1,\ldots,T_{n}$. 
With dual variables $\alpha_i^t \in \mathbb{R}$ corresponding to each equality constraint, the Lagrangian function is obtained as
 \begin{eqnarray}
\mathcal{L}\left(\mathbf{L}, b_{t},\xi_i^t\right)=\frac{{C}}{2} \sum\nolimits_{t=1}^{T} \sum\nolimits_{i=1}^{m_{t}}\left(\xi_i^t\right)^{2}+\frac{1}{2}\operatorname{tr} (\mathbf{L L}^{\top})
-\sum\nolimits_{t=1}^T\sum\nolimits_{i=1}^{m_t} \alpha_{i}^{t}( y_{i}^{t}( \left(\mathbf{L}\mathbf{u}_t\right)^{\top} \phi(\mathbf{x}_{i}^{t})+b_{t})-1+\xi_i^t),\nonumber
\end{eqnarray}
with $\mathbf{u}_{t} \triangleq [u_{t,1}, \ldots, u_{t, K}]^{\top}\in\mathbb{R}^{K}$. Then, stationary point conditions are obtained as
\begin{eqnarray}\label{eq:kkt:l}
\frac{\partial \mathcal{L}}{\partial \mathbf{L}}=0 &\Longrightarrow& \mathbf{L}=\sum\nolimits_{t=1}^T \sum\nolimits_{i=1}^{m_t} \alpha_{i}^{t}y_{i}^{t}  \phi(\mathbf{x}_{i}^{t}) \mathbf{u}_{t}^{\top},\nonumber \\
\frac{\partial \mathcal{L}}{\partial \mathbf{b}}=0 &\Longrightarrow & \mathbf{A}^{\top}\mathbf{Y} \bm \alpha =0,   \ \mathbf{b} = \left[b_{1}, \ldots, b_{t}\right]^{\top}, \nonumber\\ 
\frac{\partial \mathcal{L}}{\partial \mathbf{e}}=0 &\Longrightarrow & C \mathbf{e}=\boldsymbol{\alpha}, \nonumber\\ 
\frac{\partial \mathcal{L}}{\partial \boldsymbol{\alpha}}=0 &\Longrightarrow& \mathbf{Y}\Phi {\mathbf{w}}+ \mathbf{Y}\mathbf{A}\mathbf{b}=\mathbf{1}-\mathbf{e}.\nonumber
 \end{eqnarray}
 {where $\mathbf{A}=\text{blockdiag}(\mathbf{1}_{m_1},\cdots,\mathbf{1}_{m_T})\in\mathbb{R}^{m\times T}$, $\mathbf{w}=[(\mathbf{L}\mathbf{u}_1)^{\top},\cdots,(\mathbf{L}\mathbf{u}_T)^{\top}]^{\top}\in\mathbb{R}^{Td_h}$,  the task-specific feature mapping matrix $\Phi^t=[\phi(x_1^t),\ldots,\phi(x_{m_t}^t)]^{\top}\in\mathbb{R}^{m_t\times d_h}$  and
   $\Phi=\text{blockdiag}(\Phi^1,\cdots,\Phi^T)\in\mathbb{R}^{m\times Td_h}$  for all $T$ tasks.  All outputs, regression errors, and dual variables  are denoted as  $\mathbf{Y}=\text{diag}(y_1^1,y_2^1,\ldots,y_{m_T}^T)^{\top}\in\mathbb{R}^{m\times m}$, 
 $\mathbf{\xi}=[\xi_1^1,\xi_2^1,\ldots,\xi_{m_T}^T]^{\top}\in\mathbb{R}^{m}$, and
 ${\boldsymbol{\alpha}}=[\alpha_1^1,\alpha_2^1,\ldots,\alpha_{m_T}^T]^{\top}\in\mathbb{R}^{m}$, respectively.}

By  eliminating $\mathbf{L}$ and $\xi^t_i$, a linear system is attained as: 
\begin{equation}
    \left[\begin{array}{c|c}\mathbf{0}_{T \times T} & \mathbf{A}^{\top}\mathbf{Y}\\ \hline \mathbf{Y}\mathbf{A} & \mathbf{\mathbf { Q }}+\frac{1}{C} \mathbf{I}_{m\times m}\end{array}\right]\left[\begin{array}{c} \mathbf{b}\\ \boldsymbol{\alpha}\end{array}\right]=\left[\begin{array}{c}\mathbf{0}_T\\ \mathbf{1}_{m}\end{array}\right],
    \label{eq: update_L}
\end{equation}
where  
$\mathbf{Q}\in\mathbb{R}^{m\times m}$ is   computed by the components in  tensor $\mathcal{W}$ and the kernel function $k: \mathbb R^d \times \mathbb R^d \mapsto \mathbb R$ {induced by  $\phi(\cdot)$}, such that $\mathbf{Q}(j,j')=y_i^t y_p^q\left\langle\mathbf{u}_{t}, \mathbf{u}_{q}\right\rangle  k\left(\mathbf{x}_{i}^{t}, \mathbf{x}_{p}^{q}\right)$, 
$j=\sum_{r=1}^{t-1} m_r +i, j'=\sum_{r=1}^{q-1} m_r +p, i=1,\cdots,m_t, p=1,\cdots,m_q$ with  $i, p$ indexing the  samples in the involved tasks $t$ and $q$, respectively. With the solution of  dual variables (\ref{eq: update_L}), i.e., $\tilde{\boldsymbol{\alpha}}$,
we can get the updated  $\mathbf{L}=\sum_{t=1}^T \sum_{i=1}^{m_t} \tilde{\alpha}_{i}^{t}  y_i^t\phi(\mathbf{x}_{i}^{t}) \mathbf{u}_{t}^{\top}$. 
\vspace{0.2cm}
\noindent 2) {\bf{$\mathbf{U}^{n},b_t,\xi_i^t$-subproblem}.}
With fixed $\mathbf L$, we alternate to optimize $\mathbf{U}^{n},b^t,\xi_i^t$. 
The corresponding primal problem is:
\begin{eqnarray}
\underset{ \mathbf{u}^{n}_{t_n,:}, b_{t}, \xi_i^t}{\min }  &\frac{{C}}{2} \sum\nolimits_{t\in\mathbb{O}_{t_n}} \sum\nolimits_{i=1}^{m_{t}}\left(\xi_i^t\right)^{2}+ \frac{1}{2}\lVert \mathbf{u}_{t_{n},:}^{n}\lVert_{2}^2  \\ 
\text { s.~t. }  & y_{i}^{t}({\mathbf{u}_{t_{n},:}^{n} }^{\top}\mathbf{z}_{i}^{t} +b_{t})=1-\xi_i^t,  \nonumber
\end{eqnarray}
where 
$\mathbf{z}_{i}^{t}$ is  calculated by $\mathbf{L}^{\top}  {\phi(\mathbf{x}_{i}^{t}) } \odot \mathbf{u}_{t_{1},:}^1\odot\cdots \mathbf{u}_{t_{n-1},:}^{n-1}\odot\mathbf{u}_{t_{n+1},:}^{n+1}\odot\cdots \mathbf{u}_{t_{N},:}^N\in\mathbb{R}^K$, the involved tasks $t$ is contained in the index set
$\mathbb{O}_{t_n}=\{\overline{t_{1}, \ldots, t_{N}}| t_{l}=1,\ldots,T_{l}, l=1, \ldots, N, l \neq n\}$ with cardinality
$|\mathbb{O}_{t_n}|=\prod_{l,l\neq n}T_l$.
With dual variables $\boldsymbol{\lambda}_{t_n}$, we have the Lagrangian function:
\begin{eqnarray}
\mathcal{L}\left(\mathbf{u}^{n}_{t_{n},:}, b_{t}, {e}_i^t\right)=  \frac{{C}}{2} \sum\nolimits_{t\in\mathbb{O}_{t_n}} \sum\nolimits_{i=1}^{m_{t}}\left(\xi_i^t\right)^{2}+ \frac{1}{2}\lVert \mathbf{u}_{t_{n},:}^{n}\lVert_{2} ^2
-\sum\nolimits_{t\in\mathbb{O}_{t_n}} \sum\nolimits_{i=1}^{m_{t}}\lambda_i^t\left( y_i^t\left(  {\mathbf{u}_{t_{n},:}^{n} }^{\top}\mathbf{z}_{i}^{t} +b_{t}\right)-1+\xi_i^t \right),  \nonumber
\end{eqnarray}
where $\boldsymbol{\lambda}_{t_n} = \{\lambda^t_{i}|t\in \mathbb S_{t_n}, i=1, \ldots, m_t\}\in \mathbb{R}^{M_{t_n}}$ corresponds to  the involved constraints  in optimizing $\mathbf{u}^{n}_{t_{n},:}$.

Similarly, by deriving  the stationary  conditions 
Finding the stationary points we get 

\begin{eqnarray}
\frac{\partial \mathcal{L}}{\partial \mathbf{u}^{n}_{t_{n},:}}=0 &\Longrightarrow& \mathbf{u}^{n}_{t_{n}}=\sum_{t\in\mathbb{O}_{t_n}} \sum_{i} \lambda_{i}^{t} y_{i}^{t} \mathbf{z}_{i}^{t} \nonumber\\ 
\frac{\partial \mathcal{L}}{\partial\mathbf{b}}=0 &\Longrightarrow & \mathbf{A}_{t_n}^{\top} \mathbf{Y}_{t_n} \boldsymbol{\lambda}_{t_n}=0 \nonumber\\ 
\frac{\partial \mathcal{L}}{\partial \mathbf{e}}=0 &\Longrightarrow & C \mathbf{e}_{t_n}=\boldsymbol{\lambda}_{t_n}\nonumber\\ 
\frac{\partial \mathcal{L}}{\partial \boldsymbol{\alpha}}=0 &\Longrightarrow& \mathbf{Y}_{t_n}(\Phi_{t_n} {\mathbf{u}_{t_{n},:}^{n}}+\mathbf{A}_{t_n}\mathbf{b}_{t_n})=\mathbf{1}_{M_{ t_n}}-\mathbf{e}_{t_n}\nonumber
\end{eqnarray}
and eliminating $\mathbf{u}_{t_n,:}^n$ and $e^t_i$ therein, we get the  linear system: 
\begin{equation}
    \left[\begin{array}{c|c}
    \mathbf{0}_{\lvert \mathbb{O}_{t_n}\lvert \times \lvert \mathbb{O}_{t_n}\lvert} & \mathbf{A}_{t_n}^{\top}\mathbf{Y}_{t_n}\\ 
      \hline \mathbf{Y}_{t_n}\mathbf{A}_{t_n} & \mathbf{\mathbf { Q_{t_n} }}+\frac{1}{C} \mathbf{I}_{M_{t_n}}\end{array}\right]\left[\begin{array}{c}\mathbf{b}_{t_n}\\
    \boldsymbol{\lambda}_{t_n}\end{array}\right]=\left[\begin{array}{c}\mathbf{0}_{\lvert \mathbb{O}_{t_n}\lvert}\\ \mathbf{1}_{M_{t_n}}\end{array}\right],
    \label{eq: update_U}
\end{equation}
where $\mathbf{A}_{t_n}=\text{blockdiag}(\mathbf{1}_{m_t})\in\mathbb{R}^{M_{t_n}\times \lvert \mathbb{O}_{t_n}\lvert}$ with $t\in\mathbb{O}_{t_n}$, and ${\mathbf{Y}}_{t_n}$,  ${\boldsymbol{\lambda}}_{t_n}, \mathbf{b}_{t_n} \in \mathbb R^{M_{t_n}}$ are matrices or vectors collecting  $y_i^t$, ${\alpha}_i^t$, and $b_i^t$ involved in the equality constraints, respectively.
Here, the  matrix 
$\mathbf{Q}_{t_n}\in\mathbb{R}^{M_{t_n}\times M_{t_n}}$ is computed by $\mathbf{Q}_{t_n}(j,j')=y_i^ty_p^q\left\langle\mathbf{z}^{t}_i, \mathbf{z}_{p}^q\right\rangle $, 
where $ t, q\in\mathbb{O}_{t_n}, i=1, \ldots, m_{t},p=1,\cdots,m_q$.




\quad\quad

\begin{algorithm}
	\caption{MTL-LSSTM}
	\label{algorithm:MTL-LSSTM}
	\begin{algorithmic}
		\STATE \textbf{Input}: ${y^t_i},x^t_i$ for $t=1,\cdots,T$ $i=1,\cdots,m_t$
		\STATE \textbf{Output}: the parameters $\mathbf{U}^n$ for $n=1,\cdots,N$, $b_t$ and dual variables $\alpha_i^t$ for $t=1,\cdots,T$, $i=1,\cdots,m_t$.
		\STATE initialise $\mathbf{U}^n$ for $n=1,\cdots,N$
		\WHILE{\textbf{not convergence}}

		\STATE update ${\alpha_i^t}, b_t$ by solving  the linear system (\ref{eq: update_L})

		\FOR{$n=1:N$}
		    \FOR{$t_n=1:T_n$}
		         \STATE update ${\lambda_i^t}, b_t, t\in\mathbb{O}_{t_n}$ by solving  the linear system (\ref{eq: update_U})
		         		\STATE $\mathbf{u}^n_{t_n}=\sum_{t\in\mathbb{O}_{t_n}} \sum_{i} \lambda_{i}^{t} y_{i}^{t} \mathbf{z}_{i}^{t}$
		    \ENDFOR
		\ENDFOR
		\ENDWHILE
	\end{algorithmic}
\end{algorithm}

 \subsection{Regression}
 Please see details in "Liu, J., Tao, Q., Zhu, C., Liu, Y., \& Suykens, J. A. (2023, June). Tensorized LSSVMS For Multitask Regression. In ICASSP 2023-2023 IEEE International Conference on Acoustics, Speech and Signal Processing (ICASSP) (pp. 1-5). IEEE."

\end{document}